\documentclass[letterpaper, preprint, paper,10pt]{AAS}	
\newtheorem{problem}{Problem}

\usepackage{graphicx}
\usepackage{amssymb}
\usepackage{upgreek}
\usepackage{color}
\usepackage{amsmath}
\usepackage[tight,footnotesize]{subfigure}
\usepackage{float}
\usepackage{multirow}
\usepackage{booktabs}
\usepackage{subeqnarray}
\usepackage{epstopdf}
\usepackage{wrapfig}
\usepackage{soul}
\usepackage{siunitx}
\usepackage{longtable,tabularx}
\usepackage{xfrac}
\usepackage{overcite}
\usepackage{algorithm}
\usepackage[noend]{algorithmic}
\usepackage{xcolor}
\usepackage{subcaption}
\usepackage{subfigmat}

\usepackage{tikz}
\usetikzlibrary{shapes,arrows,positioning,calc}
\usetikzlibrary{fit,backgrounds}
\usepackage{pgfplots}
\pgfplotsset{compat=1.18}
\usepackage{pgfplotstable}
\usepackage{caption}
\tikzset{
	block/.style = {draw, fill=white, rectangle, minimum height=3em, minimum width=4em},
	alertblock/.style = {draw, color = red, 
		fill=white, rectangle, minimum height=3em, minimum width=4em},
	tmp/.style  = {coordinate}, 
	sum/.style= {draw, fill=white, circle, node distance=1cm},
	input/.style = {coordinate},
	output/.style= {coordinate},
	pinstyle/.style = {pin edge={to-,thin,black}}
}

\usepackage{setspace}
\usetikzlibrary{arrows.meta, calc, positioning, fit, backgrounds, shapes.geometric}

\newcommand{\msub}[1]{_\mathrm{#1}}                                     

\renewcommand{\leq}{\leqslant}                                          
\renewcommand{\geq}{\geqslant}                                          

\def\engE{\textsc{~e}}

\newcommand{\mydef}[1]{{\textit{#1}}}

\newcommand{\eqnnt}[1]{\hyperref[#1]{(\ref*{#1})}}
\newcommand{\eqnsnt}[2]{\hyperref[#1]{(\ref*{#1})}
	and~\hyperref[#2]{(\ref*{#2})}}
\newcommand{\eqnsernt}[2]{\hyperref[#1]{(\ref*{#1})}--\hyperref[#2]{(\ref*{#2})}}

\newcommand{\eqn}[1]{\hyperref[#1]{Eqn.~(\ref*{#1})}}
\newcommand{\eqns}[2]{\hyperref[#1]{Eqns.~(\ref*{#1})} and~\hyperref[#2]{(\ref*{#2})}}
\newcommand{\eqnser}[2]{\hyperref[#1]{Eqns.~(\ref*{#1})}--\hyperref[#2]{(\ref*{#2})}}
\newcommand{\eqnf}[1]{\hyperref[#1]{Equation~(\ref*{#1})}}
\newcommand{\eqnfs}[2]{\hyperref[#1]{Equations~(\ref*{#1})} and~\hyperref[#2]{(\ref*{#2})}}

\newcommand{\scn}[1]{\hyperref[#1]{Sec.~\ref*{#1}}}
\newcommand{\scns}[2]{\hyperref[#1]{Secs.~\ref*{#1}} and~\hyperref[#2]{\ref*{#2}}}
\newcommand{\scnser}[2]{\hyperref[#1]{Secs~.\ref*{#1}}--\hyperref[#2]{\ref*{#2}}}

\newcommand{\fig}[1]{\hyperref[#1]{Fig.~\ref*{#1}}}
\newcommand{\figs}[2]{\hyperref[#1]{Figs.~\ref*{#1}} and~\hyperref[#2]{\ref*{#2}}}
\newcommand{\figser}[2]{\hyperref[#1]{Figs.~\ref*{#1}}--\hyperref[#2]{\ref*{#2}}}
\newcommand{\figf}[1]{\hyperref[#1]{Figure~\ref*{#1}}}
\newcommand{\figfs}[2]{\hyperref[#1]{Figures~\ref*{#1}} and~\hyperref[#2]{\ref*{#2}}}
\newcommand{\figfser}[2]{\hyperref[#1]{Figures~\ref*{#1}}--\hyperref[#2]{\ref*{#2}}}

\newcommand{\tbl}[1]{\hyperref[#1]{Table~\ref*{#1}}}
\newcommand{\tbls}[2]{\hyperref[#1]{Tables~\ref*{#1}} and~\hyperref[#2]{\ref*{#2}}}
\newcommand{\tblser}[2]{\hyperref[#1]{Tables~\ref*{#1}}--\hyperref[#2]{\ref*{#2}}}

\newcommand{\apx}[1]{\hyperref[#1]{Appendix~\ref*{#1}}}

\newcommand{\prb}[1]{\hyperref[#1]{Problem~\ref*{#1}}}
\newcommand{\prp}[1]{\hyperref[#1]{Prop.~\ref*{#1}}}
\newcommand{\prpf}[1]{\hyperref[#1]{Proposition~\ref*{#1}}}

\newcommand{\algoref}[1]{\hyperref[#1]{Algorithm~\ref*{#1}}}

\newcommand{\thmref}[1]{\hyperref[#1]{Theorem~\ref*{#1}}}
\newcommand{\thmsref}[2]{\hyperref[#1]{Theorems~\ref*{#1}} and~\hyperref[#2]{\ref*{#2}}}
\newcommand{\thmserref}[2]{\hyperref[#1]{Theorems~\ref*{#1}}--\hyperref[#2]{\ref*{#2}}}

\newcommand{\algline}[1]{\hyperref[#1]{Line~\ref*{#1}}}
\newcommand{\alglines}[2]{\hyperref[#1]{Lines~\ref*{#1}} and~\hyperref[#2]{\ref*{#2}}}
\newcommand{\alglineser}[2]{\hyperref[#1]{Lines~\ref*{#1}}--\hyperref[#2]{\ref*{#2}}}

\newcommand{\enc}{E_{\nnParamEnc}}
\newcommand{\dec}{D_{\nnParamDec}}

\def\decoder{D}

\newcommand{\nnRLParam}{\xi}

\newcommand{\nnParamEnc}{\phi}
\newcommand{\nnParamDec}{\psi}

\newcommand{\nData}{N\msub{D}}
\newcommand{\nGen}{N\msub{S}}
\newcommand{\datum}{\xi}

\newcommand{\dataset}{\Xi}
\newcommand{\datasetReal}{\Xi_{\mathrm{r}}}
\newcommand{\datasetIdeal}{\Xi_{\mathrm{i}}}
\newcommand{\datasetGen}{\tilde{\Xi}}

\newcommand{\latentVector}{z}

\def\indicator{\mathcal{I}}

\newcommand\centerofmass{%
    \tikz[radius=0.4em] {%
        \fill (0,0) -- ++(0.4em,0) arc [start angle=0,end angle=90] -- ++(0,-0.8em) arc [start angle=270, end angle=180];%
        \draw (0,0) circle;%
    }%
}

\newcommand{\state}{\mathbf{x}}
\newcommand{\yaw}{\mathbf{u}}

\newcommand{\meas}{\mathbf{y}}
\newcommand{\unet}{\mathbf{\overline{\mathbf{u}}}}
\newcommand{\xvelmax}{x_{4}^{\mathrm{max}}}
\newcommand{\yvelmax}{x_{5}^{\mathrm{max}}}

\newcommand{\yawscalar}{u}
\newcommand{\xpos}{x_1}
\newcommand{\ypos}{x_2}
\newcommand{\orient}{x_3}
\newcommand{\xvel}{x_4}
\newcommand{\yvel}{x_5}
\newcommand{\angvel}{x_6}
\newcommand{\windx}{w_x}
\newcommand{\windy}{w_y}
\newcommand{\yawx}{u_1(t)}
\newcommand{\yawy}{u_2(t)}
\newcommand{\yawz}{u_3(t)}
\newcommand{\pnoise}{\omega}
\newcommand{\mnoise}{\eta}
\newcommand{\param}{\uptheta}
\newcommand{\mi}{\mathcal{I}}
\newcommand{\latentVectorjointD}{z^{*}_{1}}
\newcommand{\latentVectorjointS}{z^{*}_{2}}
\newcommand{\mrkovset}{\mathcal{M}}
\newcommand{\mrkovstate}{\mathcal{X}}
\newcommand{\mrkovaction}{\mathcal{U}}
\newcommand{\reward}{r}
\newcommand{\transistion}{p}
\newcommand{\discount}{\gamma}
\newcommand{\policy}{\pi}
\newcommand{\mvalue}{V}
\newcommand{\statevalue}{Q}
\newcommand{\advfunc}{A}
\newcommand{\shapefunc}{s}
\title{\vspace*{-\baselineskip}
	Mitigating Data Scarcity in Spaceflight Applications \\
	for Offline Reinforcement Learning \\
	using Physics-Informed Deep Generative Models}
\author{Alex E. Ballentine\thanks{Graduate Research Assistant, Aerospace Engineering Department, Worcester Polytechnic Institute, MA.}, Nachiket U. Bapat\thanks{Graduate Research Assistant, Aerospace Engineering Department, Worcester Polytechnic Institute, MA.}, and Raghvendra V. Cowlagi\thanks{Associate Professor, Aerospace
		Engineering Department. AIAA Senior Member. Corresponding Author.} }

\begin{document}

\maketitle

\section{Abstract}
The deployment of Reinforcement Learning (RL)-based controllers on physical systems is affected by poor generalization to real-world conditions, primarily due to the simulation-to-reality (sim-to-real) gap arising from unmodeled noise and stochasticity. While exposure to diverse real-world data can mitigate this issue, such data is typically scarce due to the high cost, labor, and time required for on-field experimentation. Existing approaches, including system identification and machine learning-based data augmentation, attempt to address this challenge but rely heavily on sufficient training data, the lack of which remains a fundamental limitation. In this paper, we propose the Mutual Information–based Split Variational Autoencoder (MI-VAE), a physics-augmented generative framework for synthetic data generation in low training data regimes. We evaluate our approach on an planetary Lander problem by generating synthetic datasets using both the MI-VAE and a standard Variational Autoencoder (S-VAE), and training downstream offline RL policies. Our results show that standard offline RL methods perform poorly under limited real-world data; however, augmenting these datasets with MI-VAE-generated samples significantly improves controller performance. Furthermore, the MI-VAE outperforms standard VAEs in terms of statistical fidelity, sample diversity, and downstream policy success rates, demonstrating its effectiveness in low data conditions.
	\section{Introduction}

Reinforcement learning (RL) has emerged as a powerful framework for
controller design over many domains. RL enables the direct
optimization of long-horizon performance objectives and has been shown
to operate with limited, approximate, or noisy system
dynamics\cite{zhao_dynamic_2020}. Furthermore, RL demonstrates robust
performance in managing highly non-linear and stochastic systems where
traditional controller design becomes computationally intractable or
analytically complex~\cite{10530312}. In recent years, RL policies
have led to impressive results over multiple domains, including
robotics, finance, gaming, and natural language
processing\cite{sivamayil_systematic_2023}.

Prior studies\cite{Chai2023,THANGAVEL2024100960,aerospace12090837},
have extensively explored the effectiveness and advantages of
reinforcement learning (RL) strategies for spacecraft guidance and
control problems. Tipaldi et al\cite{TIPALDI20221} survey the
literature across a broad range of domains, including planetary
landing on celestial bodies, maneuver planning, cislunar transfer and
planetary trajectory design, attitude control and guidance,
proximity and rendezvous operations, constellation orbital control,
onboard decision-making, and rover path planning. Patnala et
al\cite{PATNALA2025} present an RL-based framework employing the
cross-entropy method for celestial conjunction detection, docking
maneuver planning, and collision avoidance. Gaudat et
al.~\cite{Gaudet} develops an RL-guided approach for optimal Mars
landing, achieving pinpoint accuracy, improved fuel efficiency, and
robustness to environmental uncertainties and unmodeled dynamics.



However, naive deployment of RL controllers on real-world dynamical
systems often results in suboptimal performance. This occurs because
forward simulations based on a nominal model often fail to capture
systemic discrepancies and random disturbances present in real
environments, leading to a \emph{simulation-to-reality gap} (or
\emph{sim-to-real gap})\cite{matas_sim--real_2018}. The sim-to-reality
gap occurs as simulators, which are often used to train RL algorithms,
do not exactly represent the environment. For example, in the case of
robotic manipulation, commonly used simulators produce accumulated
errors when comparing the predicted trajectory to a real-world robotic
system under the same control input. The main consequence of this is
the same sequence of control inputs which are successful in the
simulator produce a consistently unpredictable result on the physical
system, leading to undesired behavior\cite{collins_quantifying_2019}.
Addressing the sim-to-real gap is central to successful deployment of
control policies designed in simulation to avoid performance
degradation on the actual system\cite{zhao_sim--real_2020}.

In RL, several approaches have been proposed to mitigate the
sim-to-real gap, including dynamics randomization, domain adaptation,
and the incorporation of state history\cite{kim_bridging_2025}.
Additional strategies aimed at improving policy performance and
robustness include transfer learning, human-in-the-loop training, and
inverse reinforcement learning\cite{hassani_towards_2025}. However,
the success of these techniques fundamentally depends on the quality,
diversity, and quantity of the training data
employed\cite{da2025survey,shi2022pessimistic}.

System identification techniques address this challenge by estimating
unknown system parameters from data gathered during real-world
experiments, thereby generating a nominal model that more accurately
reflects actual system dynamics and real-world effects. However,
conducting such experiments on complex systems is typically expensive,
labor-intensive, and time-consuming, resulting in limited data
availability, which can in turn compromise the effectiveness of system
identification methods\cite{billings2013nonlinear,Jategaonkar2006}.

Another promising alternative exists in the domain of machine learning
(ML). In particular, deep generative models such as Generative
Adversarial Networks
(GANs)\cite{goodfellow2014generative,creswell2018generative},
Variational Autoencoders (VAEs)\cite{kingma2019introduction-vae}, and
their variants can generate synthetic samples that closely resemble
the statistical properties of the training
data\cite{melnik2024face,vahdat2020nvae}. By augmenting scarce
real-world datasets with high-quality synthetic data obtained post
training, these models enable more effective system identification and
reinforcement learning tasks, thereby mitigating the constraints
imposed by limited experimental data.

However, a key limitation of generative models is that these models
often perform poorly when trained on limited datasets, leading to
suboptimal or biased outputs\cite{karras2020training}. To address
low-data regimes, recent research has explored methods that
incorporate additional information beyond the data itself. For
example, Physics-Informed Neural Networks (PINNs) leverage both
observed data and underlying governing laws of the system to guide
training. By enforcing adherence to the system dynamics as a
constraint, PINNs produce outputs that not only match the observed
data distribution but also satisfy the physical laws of the
system\cite{cuomo_scientific_2022,BapatPaffenrothCowlagi2024ACC}.
Nevertheless, PINNs have several important limitations. They rely
critically on the accuracy of the assumed system dynamics, so any
unmodeled deterministic effects—such as friction, drag, actuator
delays, or structural flexibility can degrade their
effectiveness~\cite{krishnapriyan2021characterizing}.

The approach presented by Bapat et
al\cite{bapat2025case,bapat2025synthetic} addresses the limited-data
challenge for time series problems through explicit latent-space
disentanglement and a strategical weighing of the``real-world" and
``simplified/simulated data" used in training. This approach enabled
the model to separately capture shared and unique features resulting
in improved training outcomes. However, the architecture did not
incorporate an explicit mechanism to regularize the interactions
between the disentangled latent spaces.

In this paper, we propose the Mutual Information–based Split
Variational Autoencoder (MI-VAE), a novel physics-informed generative
model. Briefly, this model learns the differences between system trajectories
as observed in real-world experiments and those predicted by forward
integration of physics-based equations of motion.
We generate synthetic
datasets using both the proposed MI-VAE and a standard variational
autoencoder (S-VAE), and evaluate their effectiveness in improving a
downstream RL task. We demonstrate our approach on a planetary lander
problem, focusing on two key challenges: (1) limited availability of
real-world training data, and (2) training downstream offline RL
policies using synthetic data. Our results show that standard offline
RL algorithms perform poorly when trained on limited real-world data;
however, augmenting these datasets with synthetic samples generated by
the MI-VAE significantly improves controller performance. Furthermore,
the MI-VAE outperforms standard VAEs in terms of statistical fidelity
to real-world distributions, sample diversity, and the success rate of
downstream RL policies.
Overall, this work presents a promising scalable and effective
strategy for enhancing the robustness of autonomous controllers
operating in complex, data-constrained environments.

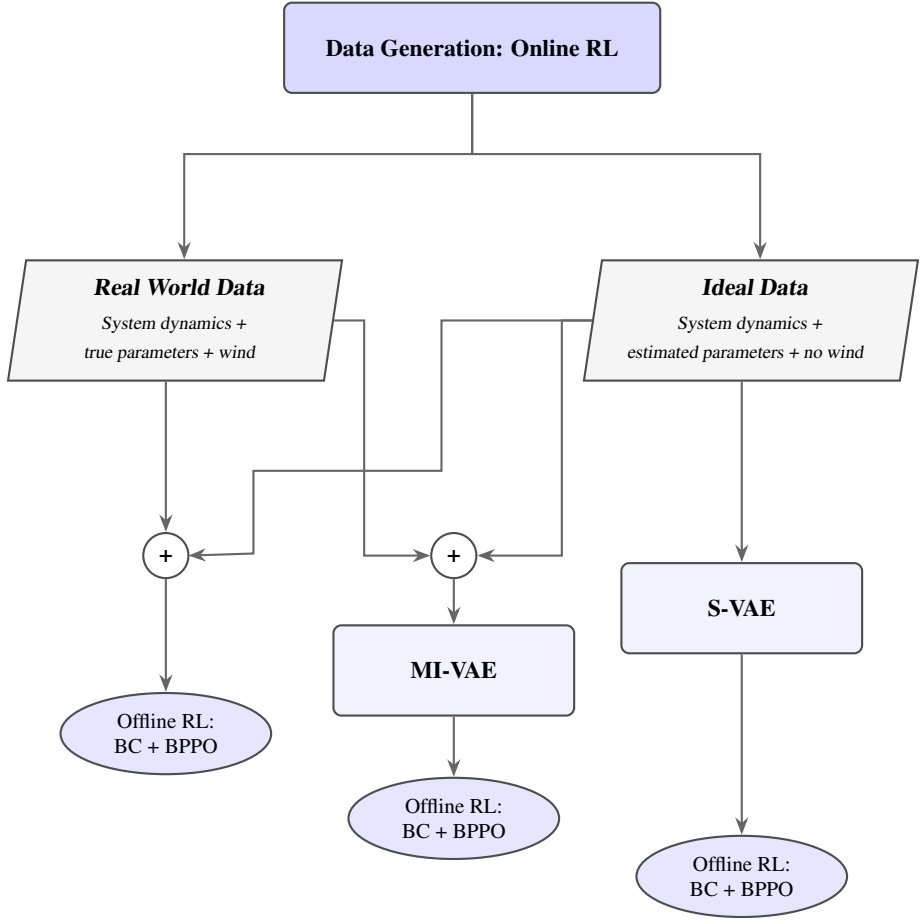
\begin{figure}[ht]
	\centering
	\begin{tikzpicture}[
		node distance=1.2cm and 0.8cm,
		base/.style = {draw=black!70, thick, align=center, font=\small},
		rect/.style = {base, rounded corners=3pt, fill=blue!5, minimum width=3.2cm, minimum height=1.2cm},
		data_shape/.style = {base, rectangle, xslant=0.15, fill=gray!8, minimum width=4.2cm, minimum height=1.6cm, font=\small\rmfamily},
		circ/.style = {draw=black!70, thick, circle, fill=white, minimum size=0.6cm, font=\small\bfseries},
		oval/.style = {base, ellipse, fill=blue!10, minimum width=2.8cm, minimum height=1cm, font=\footnotesize},
		arrow/.style = {-{Stealth[scale=1.0]}, thick, draw=black!60}
		]
		
		\node (gen) [rect, fill=blue!15, minimum width=5cm, minimum height=1.2cm] {\textbf{Data Generation: Online RL}};
		
		\node (real) [data_shape, below left=2.2cm and -0.8cm of gen] {
			\textbf{Real World Data}\\[2pt]
			\scriptsize \textup{System dynamics +}\\ \scriptsize \textup{true parameters + wind}
		};
		
		\node (ideal) [data_shape, below right=2.2cm and -0.8cm of gen] {
			\textbf{Ideal Data}\\[2pt]
			\scriptsize \textup{System dynamics +}\\ \scriptsize \textup{estimated parameters + no wind}
		};
		
		\node (plus1) [circ, below=2.0cm of real] {+};
		
		\coordinate (mid) at ($(real.south)!0.5!(ideal.south)$);
		\node (plus2) [circ, below=2.0cm of mid] {+};
		\node (mivae) [rect, below=0.6cm of plus2] {\textbf{MI-VAE}};
		
		\node (svae) [rect, below=2.4cm of ideal] {\textbf{S-VAE}};
		
		\node (out1) [oval, below=1.5cm of plus1] {Offline RL:\\BC + BPPO};
		\node (out2) [oval, below=0.8cm of mivae] {Offline RL:\\BC + BPPO};
		\node (out3) [oval, below=2.4cm of svae] {Offline RL:\\BC + BPPO};
		
		\draw [arrow] (gen.south) -- ++(0,-0.8) -| (real.north);
		\draw [arrow] (gen.south) -- ++(0,-0.8) -| (ideal.north);
		\draw [arrow] (real.south) -- (plus1.north);
		
		\draw [thick, draw=black!60] (ideal.west) -- ++(-2,0) -- ++(0,-2) -- ++(-2.5,0) -- ++(0,-1.1) 
		[arrow] -- ([xshift=0.3cm]plus1.center);
		
		\draw [arrow] (real.east) -- ++(0.4,0) |- (plus2.west);
		\draw [arrow] (ideal.west) -- ++(-0.4,0) |- (plus2.east);
		\draw [arrow] (plus1.south) -- (out1.north);
		\draw [arrow] (ideal.south) -- (svae.north);
		\draw [arrow] (svae.south) -- (out3.north);
		\draw [arrow] (plus2.south) -- (mivae.north);
		\draw [arrow] (mivae.south) -- (out2.north);
		
	\end{tikzpicture}
	\caption{Schematic overview of the overall working pipeline}
\end{figure}

\section{Problem Statement}
\label{problem-statement}
Consider a dynamical system modeled in the standard state--space form
\begin{equation}
	\dot{\state}(t) = f(\state(t), \yaw(t), \pnoise(t); \param),
	\label{eq:dyn}
\end{equation}
\begin{equation}
	\meas(t) = h(\state(t), \mnoise(t); \param),
	\label{eq:mea}
\end{equation}
 where $\state(t) \in \mathbb{R}^n$ denotes the system state, $\yaw(t) \in \mathbb{R}^p$ the control input, $\pnoise(t)$ the process noise, $\mnoise(t)$ the measurement noise, and $\param \in \mathbb{R}^m$ a vector of model parameters.  
The function $f : \mathbb{R}^n \times \mathbb{R}^p \times \mathbb{R}^{q}\rightarrow \mathbb{R}^n$ is assumed to be Lipschitz continuous in $\state$ to guarantee existence and uniqueness of solutions to \eqref{eq:dyn}.
For a fixed parameter vector $\param$, control input $\yaw(t)$, process noise realization $\pnoise(t)$, and initial condition $\state_0$ over a finite time interval $[0,T]$, a model trajectory is a function $\state : [0,T] \rightarrow \mathbb{R}^n$ that satisfies \eqref{eq:dyn}.

An \mydef{observed trajectory} of the system is an output signal $\meas(t) \in \mathbb{R}^\ell$ measured during the \mydef{real-world} operation of the system. 
In absence of process and measurement noise the \mydef{ideal} system model is  
\begin{equation}
	\dot{\state}(t) = f(\state(t), \yaw(t); \param),
	\label{eq:dyn_i}
\end{equation}
\begin{equation}
	\meas(t) = h(\state(t); \param),
	\label{eq:mea_i}
\end{equation}
The differences between the observed trajectory in ~\eqref{eq:mea} and \eqref{eq:mea_i} are a result of the real-world behavior of the system differing from ideal model dynamics due to these noise effects, namely, the sim-to-real gap.

Consider a finite sequence of time samples $t_1 < t_2 < \cdots < t_K$ within the interval $[0, T]$. 
A \mydef{datum}, or “data point,” $\datum$ consists of the output values of an observed trajectory discretized at the aforesaid 
time samples and appended with the control input $\unet =\big(\yaw(t_1),\,\yaw(t_2),\,\ldots,\,\yaw(t_K) \big) $ and the parameter value $\param$ at which the system is operated, namely, $\datum = \big( \meas(t_1),\, \meas(t_2),\, \ldots,\, \meas(t_K),\,\unet,\, \param \big) \in \mathbb{R}^{N_f},$
where $N_f := (\ell+p) K + m$. 

A general training dataset(OTD) is defined as $\dataset = \{ \datum_i \}_{i=1}^{\nData}$, where each $\datum_i$ is obtained either from \eqref{eq:dyn} and \eqref{eq:mea} yielding the real-world OTD, $\datasetReal$, or from \eqref{eq:dyn_i} and \eqref{eq:mea_i} yielding the ideal OTD, $\datasetIdeal$.


The problem of interest can then formulated as:
\begin{problem}\label{pb1}
Determine the relative performance of an offline RL algorithm trained on $\datasetReal$ versus one trained on $\datasetReal \cup \datasetGen$, where $\datasetGen = \{ \datum_j \}_{j=1}^{\nGen}$ is the synthetic dataset generated by the generative models.
\end{problem}
Implicit in this problem statement is the requirement that generating samples in $\datasetGen$ should be 
computationally efficient, so that $\nGen \gg \nData$ can be made as large as necessary.

\subsection{Planetary Lander Dynamics}
\begin{figure}[t]
\centering
\begin{tikzpicture}[scale=2, every node/.style={scale=1}]

\begin{scope}[rotate=25]
\coordinate (COM) at (0,0); 
\def\vehicleWidth{1.5}  
\def\vehicleHeight{1.5} 

\draw[thick, fill=gray!5] 
    ($(COM)+(-\vehicleWidth/2,-\vehicleHeight/2)$) rectangle 
    ($(COM)+(\vehicleWidth/2,\vehicleHeight/2)$);

\filldraw[white] (COM) circle (1pt) node[above right] { };

\draw[->, thick, black] (COM) -- ++(-0.25, 0.25) node[left] {$\mathbf{F}_d$};

\draw[->, thick, green!60!black] (COM) -- ++(0.5,-0.5) node[below] {$\mathbf{v}$};

\draw[<-, black, thick] ($(COM)+(-0.75,0.35)$) --++(-0.2,0) node[left] {$\yawz$};
\draw[<-, black, thick] ($(COM)+(0.75,-0.35)$) -- ++(0.2,0) node[right] {$\yawy$};
\draw[<-, black, thick] ($(COM)+(0.75,0.35)$) -- ++(0.2,0) node[right] {$\yawz$};
\draw[<-, black, thick] ($(COM)+(-0.75,-0.35)$) -- ++(-0.2,0) node[left] {$\yawy$};
\draw[<-, black, thick] ($(COM)+(0,-0.75)$) -- ++(0,-0.3) node[below] {$\yawx$};

\draw[black, dotted, thick] (COM) -- ++(0.5, 0) node[below] {};

\end{scope}

\draw[->, blue] (COM) ++(0.4,0) arc[start angle=0, end angle=25, radius=0.4];
\node[blue,right] at ($(COM)+(0.4,0.1)$) {$\orient$};

\draw[->, thick, black] (COM) -- ++(0,-0.4) node[below] {$\mathbf{F}_g$};

   \draw[->, thick] (-2,-1.5) -- (-1.75,-1.5) node[right] {$\xpos$};
    \draw[->, thick] (-2, -1.5) -- (-2, -1.25) node[above] {$\ypos$};

\node[shift={(0, 0)}] at (COM) {\centerofmass};

\end{tikzpicture}
\caption{Forces acting on the Mars Lander vehicle}
\label{fig:mars-lander}
\end{figure}
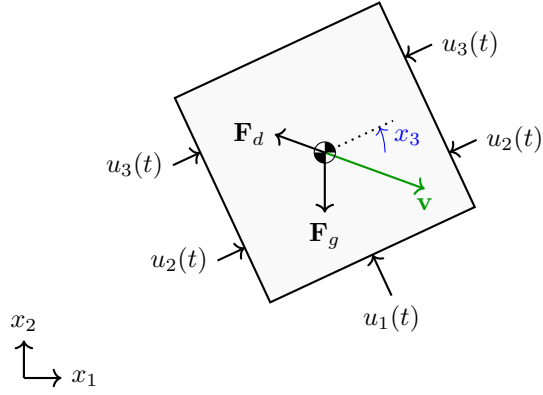

We consider a planar Lander with the following equations of motion~\eqref{eq:x-dot}-\eqref{eq:omega-dot}, which may be derived from the straight forward application of Newtonian mechanics. A quadratic drag model is included to account for aerodynamic resistance, where the drag force acts opposite to the velocity and is proportional to the square of the speed in each direction.
\begin{align}
\dot{\xpos}(t) &= \xvel(t) + \windx, \label{eq:x-dot}\\
\dot{\ypos}(t) &= \yvel(t) + \windy, \label{eq:y-dot}\\
\dot{\orient}(t) &= \angvel(t), \label{eq:theta-dot}\\
\dot{\xvel}(t) &= \frac{1}{{m}} \bigl(  -\yawx \sin \orient(t) + (\yawy + \yawz)\cos \orient(t) - {c} \xvel(t) \sqrt{\xvel^2(t)+\yvel^2(t)}\bigl),\label{eq:vx-dot}\\ 
\dot{\yvel}(t) &= \frac{1}{{m}}\bigl(-{mg} + \yawx\cos \orient(t) + (\yawy + \yawz)\sin \orient(t) - {c} \yvel(t)\sqrt{\xvel^2(t)+\yvel^2(t)}\bigl), \label{eq:vy-dot}\\ 
\dot{\angvel}(t) &= \frac{3}{2{m\ell}}(\yawy - \yawz),  \label{eq:omega-dot}
\end{align}
where $(\xpos,\ypos)$ are the position coordinates in the inertial coordinate axis system $\orient$ is the angular orientation, $(\xvel,\yvel)$ are the velocity coordinates relative to the wind, and $\angvel$ is the angular velocity. ${m}$, ${g}$, $\ell$, and ${c}$ are the vehicle mass, gravimetric acceleration, vehicle length, and drag coefficient, respectively. $\windx$ and $\windy$ are the wind velocity in the $\xpos$ and $\ypos$ directions, respectively.  
$\yawx$, $\yawy$ and $\yawz$ are thruster control inputs. $\yawy$ and $\yawz$ can exert positive or negative thrust, i.e. there are two thrusters on the vehicle and one or the other may be used. $\yawx$ only exerts a positive thrust. Figure~\ref{fig:mars-lander} shows the forces acting on the vehicle.

\section{Generative Model Architectures}\label{sec:generative-model}
%
%

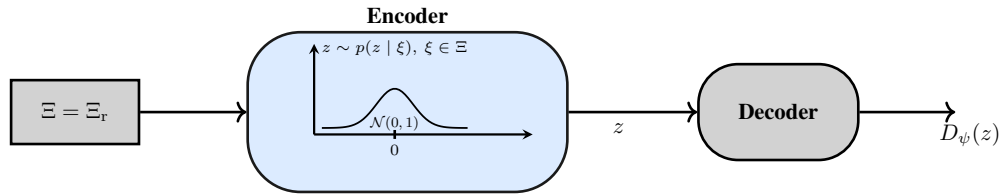
\begin{figure}[!htb]
	\centering
	\setlength{\abovecaptionskip}{-20pt}
	
	\begin{tikzpicture}[scale=0.85, transform shape]
		\definecolor{encSim}{RGB}{255,230,200}   
		\definecolor{encReal}{RGB}{220,235,255}  
		\definecolor{decCol}{RGB}{210,210,210}   
		\fill[white] (-1,-3) rectangle (12,3);
		
		\node[draw, fill=decCol, line width=1pt, text=black,
		minimum width=2cm, minimum height=1cm, inner sep=2pt] 
		(input) at (-1,0)
		{$\dataset=\datasetReal$};
		
		\node[draw=black!90!, fill = encReal,line width=1pt, rounded corners=20pt,
		minimum width=5cm, minimum height=2.5cm] 
		(encoder) at (4.2,0) {};
		\node[font=\bfseries, above] at (encoder.north) {\textbf{Encoder}};
		
		\node at (2.3,-0.35) {
			\begin{axis}[
				width=5cm,
				height=3cm,
				axis lines=left,
				line width=1pt,
				xmin=-4, xmax=4,
				ymin=0, ymax=0.6,
				xtick=\empty,
				ytick=\empty,
				clip=false
				]
			\end{axis}
		};
		
		\node at (4,0) {
			\begin{tikzpicture}[scale=0.8]
				\begin{axis}[
					width=5cm,
					height=2.5cm,
					axis lines=none,
					ytick=\empty,
					clip=false
					]
					\addplot[black, line width=1pt, domain=-4:4, samples=50]
					{exp(-x^2/2)/sqrt(2*pi)};
					\node at (axis cs:0,0.05) {\footnotesize $\mathcal{N}(0,1)$};
				\end{axis}
			\end{tikzpicture}
		};
		
		\node[draw, line width=1pt, rounded corners=15pt, fill=decCol,
		text=black, minimum width=2.5cm, minimum height=1.5cm]
		(decoder) at (10,0)
		{\textbf{Decoder}};
		
		\draw[->, line width=1pt] (input.east) -- (encoder.west);
		\draw[->, line width=1pt] (encoder.east) -- (decoder.west);
		\draw[->, line width=1pt] (decoder.east) -- ++(1.5,0);
		
		\node at (4,-0.6) {\scriptsize $0$};
		\node at (4,1)
		{\scriptsize $\latentVector \sim p(\latentVector\mid\datum),\ \datum\in\dataset$};
		\draw[line width=1pt] (4.0,-0.28)--(4.0,-0.42);
		
		\node at (7.5,-0.25) {$\latentVector$};
		\node at (13,-0.35) {$\dec(\latentVector)$};
		
	\end{tikzpicture}
	
	\caption{Architecture of a S-VAE}
	\label{fig:VAE_main}
\end{figure}

	\subsection{Variational Autoencoder Model}
A Variational Autoencoder (VAE) is a generative machine learning model that employs a probabilistic encoder--decoder framework.
For an OTD $\dataset$ and $\datum \in \dataset$, the encoder, denoted by $\enc$, performs a learnable mapping over the encoder parameters $\nnParamEnc$ as
$\enc:\mathbb{R}^n \mapsto \mathbb{R}^m$
 where $m \leq n$. The latent variable $\latentVector \in \mathbb{R}^m $ is drawn from an approximate posterior $q_{\nnParamEnc}(\latentVector|\datum)$, modeled as a Gaussian distribution
$\mathcal{N}\big(\mu(\datum; \nnParamEnc), \sigma^{2}(\datum; \nnParamEnc)\big)$.
A known prior distribution $p(\latentVector)$ is imposed on the latent representation $\latentVector$, typically chosen as $\mathcal{N}(0, 1)$.
The decoder, parameterized by $\nnParamDec$, defines the likelihood
$p_{\nnParamDec}(\datum|\latentVector)$
and aims to maximize the log-likelihood of reconstructing the observed input $\datum$ given the latent variable $\latentVector$. Figure~\ref{fig:VAE_main} illustrates the standard VAE architecture.

The Evidence Lower Bound (ELBO) is given as~\cite{kingma2019introduction-vae}
\begin{equation}
\log p_{\nnParamDec}(\datum) \ge \mathbb{E}_{\latentVector\sim q_{\nnParamEnc}(\latentVector|\datum)} \big[ \log p_{\nnParamDec}(\datum|\latentVector) \big] - D_{\mathrm{KL}}\big(q_{\nnParamEnc}(\latentVector|\datum)) \,\|\, p(\latentVector)\big)
\end{equation}
which is the log-likelihood of $\datum$ under the generative model.

 The standard VAE (S-VAE) loss is then defined as
\begin{equation}
 	\mathcal{L}_{\text{S-VAE}} = \min_{\nnParamEnc,\, \nnParamDec} 
 	\;\mathbb{E}_{\datum \in \dataset}\Big[\|\datum - D_{\nnParamDec}(\enc(\datum))\|^2 + \lambda D_{\mathrm{KL}}\Big(\mathcal{N}\big(\mu(\latentVector; \nnParamEnc), \sigma^2(\latentVector; \nnParamEnc)\big) \,\|\, \mathcal{N}(0, 1)\Big)\Big],
 \end{equation}
where the first term corresponds to the reconstruction error between the input $\datum$ and its reconstruction $D_{\nnParamDec}(\enc(\datum))$, measured using the mean squared error.  
The second term is a regularization term, given by the Kullback--Leibler divergence, which penalizes the discrepancy between the approximate posterior over the latent variables and the prior distribution, $\mathcal{N}(0, 1)$, and $\lambda$ weights this term.

It is well established that a variational autoencoder (VAE) relies on sufficiently large and diverse training data to learn a meaningful latent representation. When trained on ample data, the model can accurately capture the underlying data distribution, resulting in stable training outcomes. In contrast, when the available training data is limited, the VAE struggles to capture the true variability of the data. This often leads to issues such as overfitting, poor latent space structure, and unreliable reconstructions~\cite{kingma2019introduction-vae}. This behavior is consistent with our observations, when the S-VAE is trained on a limited subset of $\nData$ samples, its performance degrades noticeably. The corresponding results are presented in the subsequent sections.

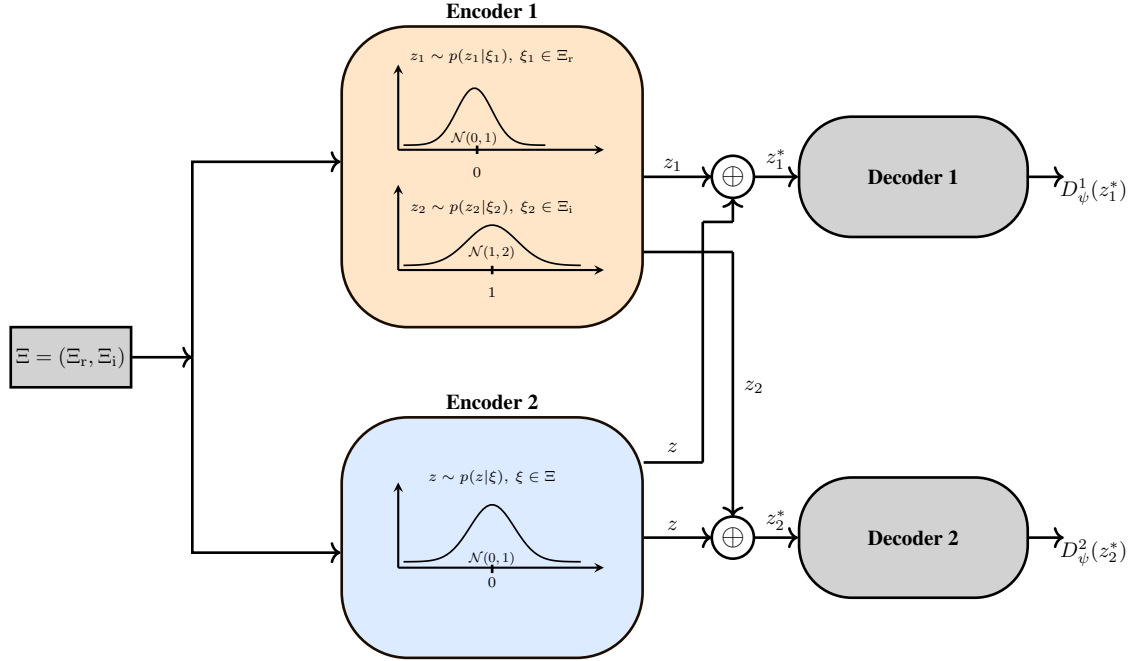
\begin{figure}[!htb]
	\centering
	\setlength{\abovecaptionskip}{20 pt} 
	\def\xshift{1} 
	
	\begin{tikzpicture}[scale=0.8, transform shape]
		\definecolor{encSim}{RGB}{255,230,200}   
		\definecolor{encReal}{RGB}{220,235,255}  
		\definecolor{decCol}{RGB}{210,210,210}   
		
		\fill[white] ({-1-\xshift},-6) rectangle ({15-\xshift},4);
		
		\node[draw,line width=1pt, fill=decCol, text=black,
		minimum width=2cm, minimum height=1cm, inner sep=2pt] 
		(input) at ({-1-\xshift},-1)
		{$\dataset =(\datasetReal,\datasetIdeal)$};
		
		\node[draw=black!90!orange,fill = encSim, line width=1pt, rounded corners=25pt,
		minimum width=5cm, minimum height=5cm] 
		(topbox) at ({6-\xshift},2) {};
		\node[font=\bfseries, above] at (topbox.north) {\textbf{Encoder 1}};
		
		\node at ({4-\xshift},2.45) {
			\begin{axis}[width=5cm,height=3cm,axis lines=left,
				line width=1pt,xmin=-4,xmax=4,ymin=0,ymax=0.6,
				xtick=\empty,ytick=\empty,clip=false]
			\end{axis}
		};
		
		\node at ({4-\xshift},0.45) {
			\begin{axis}[width=5cm,height=3cm,axis lines=left,
				line width=1pt,xmin=-4,xmax=4,ymin=0,ymax=0.6,
				xtick=\empty,ytick=\empty,clip=false]
			\end{axis}
		};
		
		\node at ({6-\xshift},2) {
			\begin{tikzpicture}[scale=0.8]
				\begin{axis}[width=6cm,height=3cm,axis lines=none,ytick=\empty,clip=false]
					\addplot[black,line width=1pt,domain=-4:4,samples=50,yshift=1.5cm]
					{exp(-x^2/2)/sqrt(2*pi)};
					\addplot[black,line width=1pt,domain=-4:6,samples=50,yshift=-1cm]
					{1/sqrt(4*pi)*exp(-(x-1)^2/4)};
					\node at (axis cs:0,0.55) {\footnotesize $\mathcal{N}(0,1)$};
					\node at (axis cs:1,-0.25) {\footnotesize $\mathcal{N}(1,2)$};
				\end{axis}
			\end{tikzpicture}
		};
		
		\draw[line width=1pt] ({10-\xshift},2) circle (0.35);
		
		\node[draw,line width=1pt,rounded corners=20pt,fill=decCol,
		text=black,minimum width=3.8cm,minimum height=2cm]
		(dec1) at ({13-\xshift},2) {\textbf{Decoder 1}};
		
		\node[draw=black!90!orange,fill = encReal, line width=1pt, rounded corners=25pt,
		minimum width=5cm, minimum height=4cm] 
		(botbox) at ({6-\xshift},-4) {};
		\node[font=\bfseries, above] at (botbox.north) {\textbf{Encoder 2}};
		
		\node at ({4-\xshift},-4.5) {
			\begin{axis}[width=5cm,height=3cm,axis lines=left,
				line width=1pt,xmin=-4,xmax=4,ymin=0,ymax=0.6,
				xtick=\empty,ytick=\empty,clip=false]
			\end{axis}
		};
		
		\node at ({6-\xshift},-4) {
			\begin{tikzpicture}[scale=0.8]
				\begin{axis}[width=6cm,height=3cm,axis lines=none,ytick=\empty,clip=false]
					\addplot[black,line width=1pt,domain=-4:4,samples=50]
					{exp(-x^2/2)/sqrt(2*pi)};
					\node[yshift=6pt] at (axis cs:0,-0.05) {\footnotesize $\mathcal{N}(0,1)$};
				\end{axis}
			\end{tikzpicture}
		};
		
		\draw[line width=1pt] ({10-\xshift},-4) circle (0.35);
		\node[font=\bfseries\Large] at ({10-\xshift},2) {$\oplus$};
		\node[font=\bfseries\Large] at ({10-\xshift},-4) {$\oplus$};
		
		\node[draw,line width=1pt,rounded corners=20pt,fill=decCol,
		text=black,minimum width=3.8cm,minimum height=2cm]
		(dec2) at ({13-\xshift},-4) {\textbf{Decoder 2}};
		
		\coordinate (split) at ($(input.east)+(1,0)$);
		\draw[->,line width=1pt] (input.east)--(split);
		\draw[->,line width=1pt] (split)|-($(topbox.west)+(0,0.25)$);
		\draw[->,line width=1pt] (split)|-($(botbox.west)+(0,-0.25)$);
		
		\draw[->,line width=1pt] (topbox.east)--({9.65-\xshift},2);
		\draw[->,line width=1pt] (botbox.east)--({9.65-\xshift},-4);
		
		\draw[line width=1pt] ($(topbox.east)+(0,-1.25)$)--({10-\xshift},0.75);
		\draw[->,line width=1pt] ({10-\xshift},0.75)--({10-\xshift},-3.65);
		\draw[line width=1pt] ($(botbox.east)+(0,1.25)$)--({9.5-\xshift},-2.75);
		\draw[line width=1pt] ({9.5-\xshift},-2.75)--({9.5-\xshift},1.25);
		\draw[line width=1pt] ({9.5-\xshift},1.25)--({10-\xshift},1.25);
		\draw[->,line width=1pt] ({10-\xshift},1.25)--({10-\xshift},1.65);
		
		\draw[->,line width=1pt] ({10.35-\xshift},2)--(dec1.west)
		node[midway,above] {$\latentVectorjointD$};
		\draw[->,line width=1pt] ({10.35-\xshift},-4)--(dec2.west)
		node[midway,above] {$\latentVectorjointS$};
		
		\draw[->,line width=1pt] (dec1.east)--({15.5-\xshift},2);
		\draw[->,line width=1pt] (dec2.east)--({15.5-\xshift},-4);
		
		\node at ({9-\xshift},2.2) {$\latentVector_1$};
		\node at ({9-\xshift},-3.8) {$\latentVector$};
		\node at ({9-\xshift},-2.5) {$\latentVector$};
		\node at ({10.35-\xshift},-1.5) {$\latentVector_2$};
		
		\node at ({6-\xshift},4)
		{\scriptsize$\latentVector_1 \sim p(\latentVector_1|\datum_1),\ \datum_1\in\datasetReal$};
		\node at ({6-\xshift},1.5)
		{\scriptsize$\latentVector_2 \sim p(\latentVector_2|\datum_2),\ \datum_2\in\datasetIdeal$};
		\node at ({6-\xshift},-3)
		{\scriptsize$\latentVector \sim p(\latentVector|\datum),\ \datum\in\dataset$};
		
		\node at ({16-\xshift},1.75) {$\dec^1(\latentVectorjointD)$};
		\node at ({16-\xshift},-4.25) {$\dec^2(\latentVectorjointS)$};
		
		\draw[line width=1pt] (5,0.38)--(5,0.52);
		\draw[line width=1pt] (4.75,2.38)--(4.75,2.52);
		\draw[line width=1pt] (5,-4.45)--(5,-4.59);
		\node at (5,0.08) {\scriptsize $1$};
		\node at (4.75,2.08) {\scriptsize $0$};
		\node at (5,-4.75) {\scriptsize $0$};
		
	\end{tikzpicture}
	
	\caption{Architecture of MI-VAE}
	\label{fig:MI-VAE_main}
\end{figure}

\subsection{Mutual Information based Split Variational Autoencoder Model}\label{MI_Split} 
	To address the issue of training the S-VAE with scarce data we propose the MI-VAE model. The proposed MI-VAE trains strategically on both the OTD and the ideal data(refer to prior sections) to mitigate the data scarcity issue.     The MI-VAE employs a probabilistic encoder--decoder framework comprising two encoders and two decoders, parameterized by the learnable set $\{\nnParamEnc_1, \nnParamEnc_2, \nnParamDec_1, \nnParamDec_2\}$.  
	The latent space of Encoder~1 (see Figure~\ref{fig:MI-VAE_main}) is split into two subspaces, each dedicated to capturing distinct aspects of the OTD. 
	The OTD appends two datasets, $\datasetReal$ and $\datasetIdeal$, of input components, i.e. $\dataset = \{\datasetReal,\datasetIdeal\}$.
	
 Encoder~1 implements the mapping
\[
E^{1}_{\nnParamEnc_1} : \mathbb{R}^n \longrightarrow \mathbb{R}^m,
\]
defined by
\[
E^{1}_{\nnParamEnc_1}(\datum) =
\begin{cases}
	q_{\nnParamEnc_1}(z_1 \mid \datum_1), & \text{if } \datum_1 \in \datasetReal,\\
	q_{\nnParamEnc_1}(z_2 \mid \datum_2), & \text{if } \datum_2 \in \datasetIdeal.
\end{cases}
\]

	The proposed posterior distributions $q_{\nnParamEnc_1}(z_1 \mid \datum)$ and $q_{\nnParamEnc_1}(z_2 \mid \datum)$ are modeled as Gaussian distributions $z_1 \sim \mathcal{N}(\mu(z_1; \nnParamEnc_1), \sigma^2(z_1; \nnParamEnc_1))$ and $z_2 \sim \mathcal{N}(\mu(z_2; \nnParamEnc_1), \sigma^2(z_2; \nnParamEnc_1))$.  
	We select two distinct priors, $p(\latentVector_1)$ and $p(\latentVector_2)$, as $\mathcal{N}(0, 1)$ and $\mathcal{N}(1, 2)$, respectively. This choice of priors is arbitrary and only serves to capture the distinct aspects of the input.
	
	A mutual information–informed loss term $\mi_{\nnParamEnc_1}(\latentVector_1, \latentVector_2 \mid \datum_1, \datum_2)$ is incorporated to encourage independence between these approximate posteriors, thereby isolating characteristics unique to each dataset within the latent space.

	Encoder~2 (also shown in Figure~\ref{fig:MI-VAE_main}) maps any $\datum$ to a common latent representation by performing the encoding operation given by $E^{2}_{\nnParamEnc_2}: \mathbb{R}^n \mapsto \mathbb{R}^m$. The latent variable $\latentVector$  follows the proposed posterior $q_{\nnParamEnc_2}(\latentVector\mid \datum)$, which is modeled as a Gaussian distribution $\mathcal{N}(\mu(z;\nnParamEnc_2), \sigma^{2}(z;\nnParamEnc_2)),$
	with the prior set as $\mathcal{N}(0, 1)$. This allows the model to capture shared characteristics between the two datasets in a unified latent space.
	
	During reconstruction, we sample $\latentVector_1$, $\latentVector_2$, and $\latentVector$, and concatenate them to form the latent vectors $\latentVectorjointD = [\latentVector_1, \latentVector]$ and $\latentVectorjointS = [\latentVector_2, \latentVector]$, which are subsequently passed through Decoder~1 and Decoder~2, respectively.  
	Decoder~1 and Decoder~2 are trained to maximize the likelihoods $p(\datum_1 \mid \latentVectorjointD)$ and $p(\datum_2 \mid \latentVectorjointS)$, thereby reconstructing the observed inputs $\datum_1$ and $\datum_2$ given $\latentVectorjointD$ and $\latentVectorjointS$.

	To that end, we use the following loss function:
	\begin{equation}
		\begin{aligned}
			\mathcal{L}_{\text{MI-VAE}} = \min_{\nnParamEnc_1, \nnParamEnc_2, \nnParamDec_1, \nnParamDec_2} 
			\; \mathbb{E}_{{\datum_1  \in \datasetReal},{\datum_2 \in \datasetIdeal}}   
			\Big\{ 
			& \,\lambda_1\,\mathrm{KL}\big(\mathcal{N}(\mu(\latentVector_1;\nnParamEnc_1), \sigma^{2}(\latentVector_1;\nnParamEnc_1))\|\,\mathcal{N}(0,1)\big)\\
			&+ \lambda_2\,\mathrm{KL}\big(\mathcal{N}(\mu(\latentVector_2;\nnParamEnc_1), \sigma^{2}(\latentVector_2;\nnParamEnc_1))\|\,\mathcal{N}(1,2)\big) \\
			&+ \,\beta\, \mi_{\nnParamEnc_1}(\latentVector_1, \latentVector_2 \mid \datum_1, \datum_2)
			+ \lambda_4\,\mathrm{KL}\big(\mathcal{N}(\mu(\latentVector;\nnParamEnc_2), \sigma^{2}(\latentVector;\nnParamEnc_2))\|\,\mathcal{N}(0,1)\big) \\
			& +\, \lVert \datum_1 - \decoder^1_{\nnParamDec_1}(\latentVectorjointD)\rVert^2
			+ \lVert \datum_2 - \decoder^2_{\nnParamDec_2}(\latentVectorjointS) \rVert^2
			\Big\}.
		\end{aligned}
	\end{equation}
	
	The reconstruction loss is computed using the mean squared error (MSE) between the reconstructed outputs and the corresponding inputs. $\decoder^1_{\nnParamDec_1}(\latentVectorjointD)$ and $\decoder^2_{\nnParamDec_2}(\latentVectorjointS)$ denote the decoding operations of Decoder 1 and Decoder 2, respectively. The $\lambda 's$ represent the weights for the KL divergences and $\beta$ the ML regularization term.

	As indicated in the results section below, the improvement in sample quality of the MI-VAE arises because $\datasetIdeal$ contains a larger and more diverse set of samples, enabled by the computational efficiency of forward-simulating system dynamics for a nominal parameter value. In contrast, $\datasetReal$ is limited due to practical constraints of real-world experiments. Consequently, we obtain more numerous and diverse latent samples corresponding to $\latentVector_2$ and $\latentVector$. By concatenating each latent code $\latentVector_1$ with all available $\latentVector$ samples and repeating this process until the full set of $\latentVector$ is exhausted, we construct a larger and more diverse set of combined latent representations. Since the number of $\latentVector_1$ samples is smaller than that of $\latentVector$, the $\latentVector_1$ samples are repeated to pair with every corresponding $\latentVector$ sample. This repetition does not introduce redundancy, as each concatenated vector $\latentVectorjointD = [\latentVector_1, \latentVector]$ remains unique due to the variability contributed by $\latentVector$. As a result, the effective latent distribution becomes substantially richer and more diverse.
	
	This enriched latent space enhances the model’s capacity to represent the underlying variability of the observed data, thereby improving generalization. In contrast, the standard VAE (S-VAE), trained solely on the limited OTD $\datasetReal$, lacks this mechanism for latent space augmentation and exhibits reduced diversity in its learned representations.

\section{Reinforcement Learning}\label{sec:rl}
We use reinforcement learning (RL) to generate realistic, controlled trajectories of the Mars Lander. RL is a machine learning paradigm that seeks to maximize cumulative rewards in a given environment through trial-and-error \cite{barto_reinforcement_1997}. In comparison with traditional optimization techniques, RL provides an easier method of generating large quantities of data. While solving optimal control equations (i.e. the Hamilton-Jacobi-Bellman equation) may require solving high-dimensional non-linear systems of equations, data generation using a trained RL policy only requires forward simulation of the environment and querying the policy at each step. Note that we consider discrete-time RL algorithms, in that the state of the environment is sampled at uniform discrete timesteps $\Delta t$. In this way, RL allows for scalable data generation of physically reasonable, controlled system at comparatively low computational costs. 

The core of RL is a sequential decision-making process, often modeled as a Markov Decision Process (MDP) $\mrkovset = \{\mrkovstate, \mrkovaction, \reward, \transistion, \discount\}$ \cite{sutton_reinforcement_2018}. $\mrkovstate$ is the state space, $\mrkovaction$ is the action or control space, $\reward$ is a scalar reward function, $\transistion$ governs the transition dynamics of the system, and $\discount$ is a discount factor on future rewards. RL algorithms aim to learn a policy $\policy(\yaw(t)|\state(t))$, which may be deterministic or a probability distribution of actions $\yaw(t) \in \mrkovaction$ conditioned on the current state of the system $\state(t) \in \mrkovstate$ \cite{ballentine_inverse_2025}. For a policy $\policy$, an associated value function $\mvalue^\policy$ may be determined for any $\state(t)$ as
\begin{equation}
\mvalue^\policy(\state(t))=\mathbb{E}_{\state(t), \policy} \biggl[\sum_{k=0}^K \discount^t \reward(\state(t+k\Delta t), \yaw(t+k\Delta t))|\state_0=\state(t) \biggl]
\end{equation}
 where $t$ is the time along the trajectory, $\state(t)$ is the initial state, and $K$ is the total number of discrete timesteps, which may be infinite. Similarly, we can define a state-action value function $\statevalue^\policy$, which is the discounted return given action $\yaw(t)$ chosen at state $\state(t)$ \cite{zhuang_behavior_2023}: 
\begin{align}
\statevalue^\policy(\state(t), \yaw(t))= \mathbb{E}_{\state(t), \yaw(t), \policy} \biggl[\sum_{k=0}^K \discount^t \reward(\state(t+k\Delta t), \yaw(t+k\Delta t))|\state_0=\state(t), \yaw_0=\yaw(t) \biggl]
\end{align}

Given the value function and state-action value function, we define the advantage function as $\advfunc^\policy(\state(t),\yaw(t))=\statevalue^\policy(\state(t),\yaw(t)) - \mvalue^\policy(\state(t))$, which gives the relative advantage of an action $\yaw(t)$ given state $\state(t)$ \cite{sutton_advances_1999}. 

\subsection{Proximal Policy Optimization}
Proximal policy optimization (PPO) is an on-policy, policy gradient method \cite{schulman_proximal_2017}. Policy gradient methods directly learn a policy function $\policy_\nnRLParam$ which maximizes the expected long-term reward. On-policy methods use data collected from the current policy, rather than using a replay buffer or prior collected transitions. While on-policy methods are less sample efficient than off-policy methods--which use a replay buffer, they can also be more robust and provide more theoretical convergence guarantees \cite{hammami_policy_2022}. PPO provides a simple, easy-to-implement reinforcement learning algorithm for both continuous and discrete action spaces, which also performs well over common RL benchmarks \cite{schulman_proximal_2017}. 

We use one of the two main variants of PPO, PPO-clip. At every update of the policy, PPO-clip restricts the amount that the policy can change, which helps improve the stability of the training process. The objective function of PPO optimized through gradient ascent and is given as follows: 
\begin{equation}\nnRLParam_{k+1}=\arg\max_\nnRLParam \mathbb{E}_{\state(t),\yaw(t) \sim\policy_{\nnRLParam_k}}[\mathcal{L}(\state(t),\yaw(t),\nnRLParam_k, \nnRLParam)]
\end{equation}
where $\mathcal{L}$ is given by:
\begin{align}
\mathcal{L}(\state(t),\yaw(t),\nnRLParam_k, \nnRLParam) & = \min\biggl( \frac{\policy_\nnRLParam(\yaw(t)|\state(t))}{\policy_{\nnRLParam_k}(\yaw(t)|\state(t))} \hat{\advfunc}(\state(t),\yaw(t)), \: g(\epsilon, \hat{\advfunc}(\state(t),\yaw(t))) \biggl) \\
g(\epsilon, \advfunc) & = \begin{cases}  
(1 + \epsilon)\advfunc &\text{if } \advfunc \geq 0 \\  
(1 - \epsilon)\advfunc &\text{if } \advfunc < 0  
\end{cases}
\end{align}
where $\epsilon$ is a hyperparameter governing the allowed divergence between the old and new policies. $\frac{\policy_\nnRLParam(\yaw(t)|\state(t))}{\policy_{\nnRLParam_k}(\yaw(t)|\state(t))}$ is the importance sampling ratio, which compares the value of the current policy to the old policy. $\hat{\advfunc}(\state(t),\yaw(t))$ is an estimate of the advantage function. $g(\epsilon, \advfunc)$ is included in the objective function to clip the maximum allowed value of $\mathcal{L}$ and avoid large, destabilizing updates. The basic principle of PPO is to increase the probability of a certain action given a state, i.e. increase $\policy_\nnRLParam(\yaw(t)|\state(t))$, if the advantage is positive. If the advantage is negative, i.e. a particular action is expected to the less advantageous than other possible actions, the probability of selecting that action is decreased. Clipping the objective function ensures that the policy is not updated too rapidly \cite{schulman_proximal_2017, schulman_high_dimensional_2018}. 

\subsection{Behavior Proximal Policy Optimization}
Assuming that we are able to sample a large number of realistic trajectories, we would like to be able to use that data and design a controller. This gathered data should represent the realistic system under a large number of conditions. To generate the initial set of trajectories, we used reinforcement learning to learn a semi-optimal controller under a given cost function. However, in this case, we assume that we have access to a large number of sample trajectories, but we may not have access to an exact simulation of the system. Therefore, we want to train some system to learn a good controller, while primarily relying on the sample data instead of an exact simulation of the system. 

We consider offline RL to solve this problem. Typically, RL algorithms are online, meaning that the agent interacts with the environment to gather data according to its current policy. However, offline RL refers to a different training process wherein the agent learns solely from a fixed dataset and has no interaction with the environment \cite{levine_offline_2020}. We use Behavior Proximal Policy Optimization (BPPO) to solve this offline RL problem. 

BPPO uses data from demonstrations to first learn an approximation of the ``expert" policy $\hat{\policy}_\beta$, the state-action value function $\hat{\statevalue}^{\policy_\beta}$, and the value function $\hat{\mvalue}^{\policy_\beta}$. The behavior policy $\hat{\policy}_\beta$ is determined using behavior cloning: 
\begin{equation}
\hat{\policy}_\beta = \arg \max_\policy \mathbb{E}_{(\state(t),\yaw(t))\sim\mathcal{D}} [\log \policy(\yaw(t)|\state(t))]
\end{equation}
 where $\hat{\policy}_\beta$ is learned using sampled state-action pairs $(\state(t),\yaw(t))$ from a fixed dataset $\mathcal{D}=\{(\state(t), \yaw(t), \state(t+\Delta t), \reward(t))\}$ of transitions \cite{zhuang_behavior_2023}. The state-action function is calculated using the SARSA algorithm, where the state-action value function is updated using the Bellman equation \cite{yao_improved_sarsa_2025}: 
 \begin{equation}
 \statevalue(\state(t), \yaw(t)) = \reward(\state(t),\yaw(t)) + \discount \statevalue(\state(t+\Delta t), \yaw(t+\Delta t))
 \end{equation}
 The value function is learned by fitting the cumulative returns for states $\state(t) \in \mathcal{D}$. 

In BPPO, the initial guess for the learned policy $\policy_{\nnRLParam}$ is the behavior cloned policy $\policy_\beta$. BPPO uses the same update rule as PPO-clip, except $\policy_{\nnRLParam_k}$ is fixed at $\policy_\beta$ and $\hat{\advfunc}$ is approximated using  $\hat{\statevalue}^{\policy_\beta}$ and $\hat{\mvalue}^{\policy_\beta}$. We also periodically calculate the expected cumulative reward of $\policy_\nnRLParam$ through interaction with the environment, although the transitions are not saved or used to update the policy. We choose to use information about the agent's performance in the environment is used to over-write $\policy_\beta$ whenever the performance outperforms the baseline behavior-cloned performance \cite{zhuang_behavior_2023}. 

\begin{table}[t]
\centering
\caption{True and estimated vehicle parameters}
\label{tab:parameter-sets}
\setlength{\tabcolsep}{4pt}  
{\fontsize{9pt}{11pt}
\begin{tabular}{lcccccccc}
\hline
\textbf{Parameter Set} & ${m}$ [$\mathrm{m}$] & ${g}$ [$\mathrm{\sfrac{m}{s^2}}$] & ${l}$ [$\mathrm{m}$]& ${c}$ & $\Delta t$ [$\mathrm{s}$]
& $\yawx^\mathrm{max}$ [$\mathrm{kN}$] & $\yawy^\mathrm{max}$ [$\mathrm{kN}$] & $\yawz^\mathrm{max}$ [$\mathrm{kN}$]\\ 
\hline
PA & $500$ & $3.728$ & $10$ & $0.2$ & $0.05$ & $15$ & $2$ & $2$ \\
PB & $450$ & $3.65$ & $11$ & $0.4$ & $0.05$ & $10$ & $5$ & $5$\\ 
\hline
\end{tabular}}
\end{table}

\section{Results and Discussion}
\label{sec-results}
In this section, we evaluate the quality of the synthetic data generated by the S-VAE and MI-VAE models. We then present results comparing BC and BPPO training outcomes when using real-world data versus synthetic data.

We implemented the S-VAE and MI-VAE generative models using PyTorch \cite{paszke2019pytorch}, 
which is a library of Python-based software tools for implementing various NN architectures. We used Stable Baselines3 \cite{stable-baselines3} to implement PPO, and we modified the reference implementation of BPPO to implement the offline RL algorithm \cite{zhuang_behavior_2023}.


\subsection{Data Generation}\label{sec:data-generation}

\begin{table}
\centering
\caption{Trajectory Characteristics}
\label{tab:state-bounds}
\setlength{\tabcolsep}{4pt}  
{\fontsize{9pt}{11pt}
\begin{tabular}{lcccccccc}
\hline

& $\xpos$ & $\ypos$ & $\orient$ & $\xvel$ & $\yvel$ & $\angvel$ & $\windx$ & $\windy$ \\
\hline
$\state_0$ Lower Bound
& $-50$ & $150$ & $-\frac{\policy}{6}$ & $-5$ & $-20$ & $-0.5$ & $-4$ & $-1$ \\

$\state_0$ Upper Bound
& $50$ & $200$ & $\frac{\policy}{6}$ & $5$ & $-2.5$ & $0.5$ & $4$ & $1$ \\

$\state_f$ Lower Bound
& $-4$ & $0$ & $-\frac{\policy}{18}$ & $-3-\windx$ & $-3-\windy$ & $-$ & $-$ & $-$ \\

$\state_f$ Upper Bound
& $4$ & $1$ & $\frac{\policy}{18}$ & $3-\windx$ & $-\windy$ & $-$ & $-$ & $-$ \\

$\state_f$ Target Value
& $0$ & $<1$ & $0$ & $-\windx$ & $-\windy$ & $-$ & $-$ & $-$ \\
\hline
\end{tabular}}
\end{table}

We synthesized 2 training datasets under distinct parameters, Parameter Set A (PA) and Parameter Set B (PB) for $\datasetReal$ and $\datasetIdeal$, respectively. Additionally, $\datasetReal$ includes wind in the system dynamics, which we consider to include both wind and uncertainties in the environment model. $\datasetIdeal$ does not include wind, i.e. $\windx$ and $\windy$ are taken to be zero.  
Table~\ref{tab:parameter-sets} gives both sets of parameters of the vehicle. Each parameter set is used to train a RL policy using PPO. The reward function, given in \eqref{eq:reward-ppo}, is the same for the two policies and is a combination of three terms: a shaping term, control penalty term, and a terminal penalty.
\begin{align}
r &= 0.5 \Big( \shapefunc^2 - \shapefunc_\mathrm{prev}^2 \Big)
     - 0.1 \sum_{i=1}^{3} \frac{u_i(t)}{u_i^\mathrm{max}} \label{eq:reward-ppo} \\
     &\quad + \indicator_{\{\ypos < 1\}} \Bigg( 100
       - 2 |\xpos|
       - \frac{180}{\policy} |\orient|
       - 5 |\xvel + \windx|
       - 10 |\yvel + \windy| \Bigg),   \nonumber \\[2mm]
\shapefunc &= \sum_{i=1}^{3} \frac{\state_i(t)}{\state_i^\mathrm{max}}
     + \frac{\xvel + \windx}{\xvelmax}
     + \frac{\yvel + \windy}{\yvelmax} \label{eq:shaping}
\end{align}
where $\shapefunc$ is the shaping term, specified in \eqref{eq:shaping} and $\shapefunc_\mathrm{prev}$ is the value of the shaping term at the prior timestep. $\shapefunc^2 - \shapefunc_\mathrm{prev}^2$ provides an incremental reward which guides the algorithm towards the intended goal point. The shaping term is known to preserve the optimal policy while providing a ``denser" reward function to improve convergence of the algorithm~\cite{ng_invariance_1999}. The second term in the reward function penalizes larger control inputs. Both the shaping term and the control penalty are scaled by a maximum value for each state to avoid disproportionately large rewards/penalties associated with a particular state or control input. Finally, the last term is a terminal state reward multiplied by $\indicator_{\{\ypos < 1\}}$, which is 1 if $\ypos<1$ and 0 otherwise. The terminal state reward gives the algorithm a positive reward of 100 for reaching a terminal state, in this case $\ypos < 1$, and penalizes any deviation from zero for $\xpos$, $\orient$, $\xvel$, and $\yvel$. The coefficients on the these terms were determined through experimentation to achieve the desired performance over all final states. 

For both training and data generation, we randomly generated initial states and wind profiles from a uniform distribution between an upper and lower bound, which is unique for all states. The bounds on the initial conditions for each state are given in Table~\ref{tab:state-bounds}. A trajectory is considered to be successful if the final state falls within the final state bounds also given in Table~\ref{tab:state-bounds}. We process each trajectory before further use by discretizing the trajectory into a fixed length of $N=100$ instead of a fixed timestep $\Delta t = 0.05$. We use one-dimensional linear interpolation for each state to interpolate the values at intermediate times not given in the initial trajectory.

\subsection{S-VAE and MI-VAE Implementation}

For both VAEs, the encoder and decoder networks were implemented as multilayer perceptrons with a latent dimensionality of 32 for each latent variable. The S-VAE and MI-VAE architectures employed four hidden layers in total, with each hidden layer consisting of 324 neurons. Rectified Linear Unit (ReLU) activations~\cite{Banerjee2019} were used throughout the networks, and Layer Normalization was applied after every hidden layer to improve training stability. The models were trained using the Adam optimizer with a learning rate of $10^{-3}$ and a batch size of $M_b = 32$. Training was performed for 2000 epochs for the S-VAE and 1000 epochs for the MI-VAE with fixed random initialization to ensure reproducibility.
The MI regularization weight for the MI-VAE was set to $\beta = 20.0$, and a warm-up period of 50 epochs was used during which the MI term is excluded to stabilize early training. Mutual information was estimated using an exponential moving average (EMA) of the joint covariance with a decay factor of 0.99. The ideal training dataset, $\datasetIdeal$, was fixed at 1000 data points for the MI-VAE. This value was selected empirically, as increasing the size of the ideal dataset beyond 1000 samples did not yield any statistically significant improvement in training performance.
All experiments were conducted using standardized input data, with feature-wise normalization applied independently to each dataset.
Note all hyperparameter combinations were obtained after a series of trials.

%

\begin{table*}[t]
\centering
\caption{Mean and standard deviation of average state deviation for $\state_1$ to $\state_6$ over 1000 generated trajectories based on predicted state and integrated state using predicted control input.}
\label{tab:trajectory-deviation}
\setlength{\tabcolsep}{4pt}  
{\fontsize{9pt}{11pt}
\begin{tabular}{l | c c c c c c }
\hline
\textbf{Model}
& \multicolumn{6}{c}{\textbf{Average State Deviation}} \\
\cmidrule{2-7}
& $\xpos$ & $\ypos$ & $\orient$ & $\xvel$ & $\yvel$ & $\angvel$ \\
\hline
S-VAE-25  & $3.20 \pm 1.70$ & $5.08 \pm 3.07$  & $0.12 \pm 0.08$  & $1.69 \pm 0.97$ & $1.60 \pm 0.81$ & $0.029 \pm 0.017$ \\ 
MI-VAE-25  & $\mathbf{1.27 \pm 1.31}$ & $\mathbf{1.95 \pm 2.22}$ & $\mathbf{0.03 \pm 0.04}$ & $\mathbf{0.59 \pm 0.51}$ & $\mathbf{0.67 \pm 0.71}$ & $\mathbf{0.008 \pm 0.008}$ \\ 
\midrule
S-VAE-1000  & $2.26 \pm 2.18$ & $2.95 \pm 3.18$ & $0.09 \pm 0.09$ & $1.49 \pm 1.37$ & $0.96 \pm 0.99$ & $0.024 \pm 0.021$ \\
MI-VAE-1000 & $\mathbf{1.72 \pm 1.54}$ & $\mathbf{2.92 \pm 3.69}$ & $\mathbf{0.07 \pm 0.06}$ & $\mathbf{1.12 \pm 0.98}$ & $\mathbf{0.88 \pm 1.05}$ & $\mathbf{0.018 \pm 0.014}$ \\
\hline
\end{tabular}}
\end{table*}

\paragraph{Visual Assessment:}
Figures~\ref{fig-vae-mars-25}-\ref{fig-MI-VAE-mars-1000} show the trajectory plotted in $\xpos$ and $\ypos$ for the S-VAE and MI-VAE models, plotted alongside their respective simulated trajectories obtained by forward propagating the system dynamics given in Equations~\eqref{eq:x-dot} and~\eqref{eq:y-dot}   using the initial conditions and drift terms as the generated trajectory. It can be seen that the trajectories generated by the MI-VAE more closely align with their numerically simulated counterparts than those produced by the S-VAE when trained on 25 samples from $\datasetReal$. This indicates the adherence to the entailing system dynamics. For 1000 $\datasetReal$ samples, the performance of both models is comparable. To quantify the deviation of the generated state trajectories from their simulated equivalents, an MAE-based performance metric is employed. Furthermore, several discontinuities or ``kinks" are observed in the S-VAE output trajectories, whereas the MI-VAE trajectories remain smooth for the 25-sample case. Table~\ref{tab:trajectory-deviation} presents the average state deviation over 1000 trajectories for the S-VAE and MI-VAE for all states in the model, with best results bolded. We see that the MI-VAE consistently produces trajectories with the lowest average deviation over all states gives both 25 and 1000 training samples from $\datasetReal$.
\def\thisfigwidth{0.46\columnwidth}
\def\thisfigspace{0.01\columnwidth}
\begin{figure}
	\centering
	\begin{subfigmatrix}{2}
		\subfigure{\includegraphics[width=\thisfigwidth]{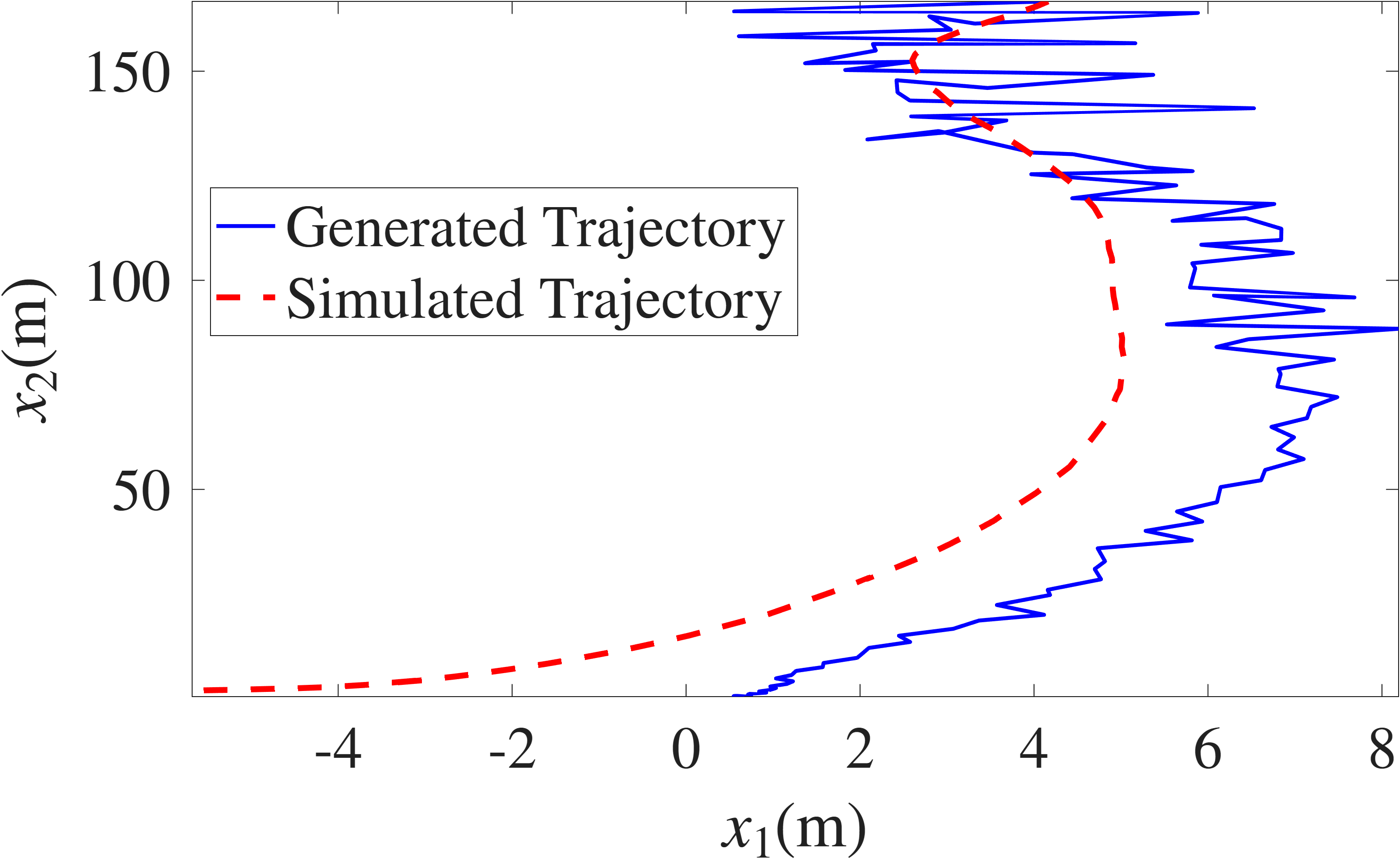}}
		\hspace{\thisfigspace}
		\subfigure{\includegraphics[width=\thisfigwidth]{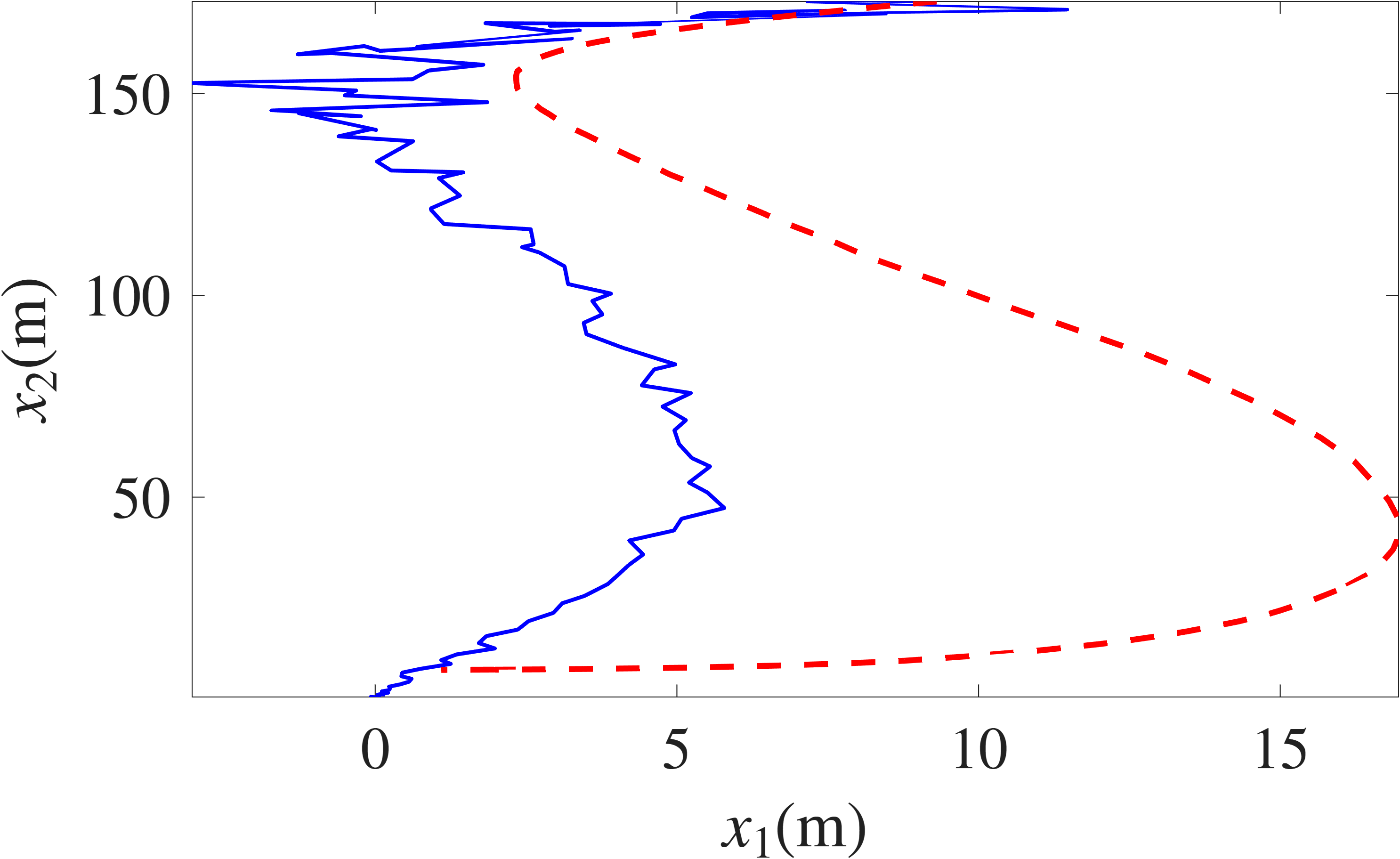}}
		\hspace{\thisfigspace}
		\subfigure{\includegraphics[width=\thisfigwidth]{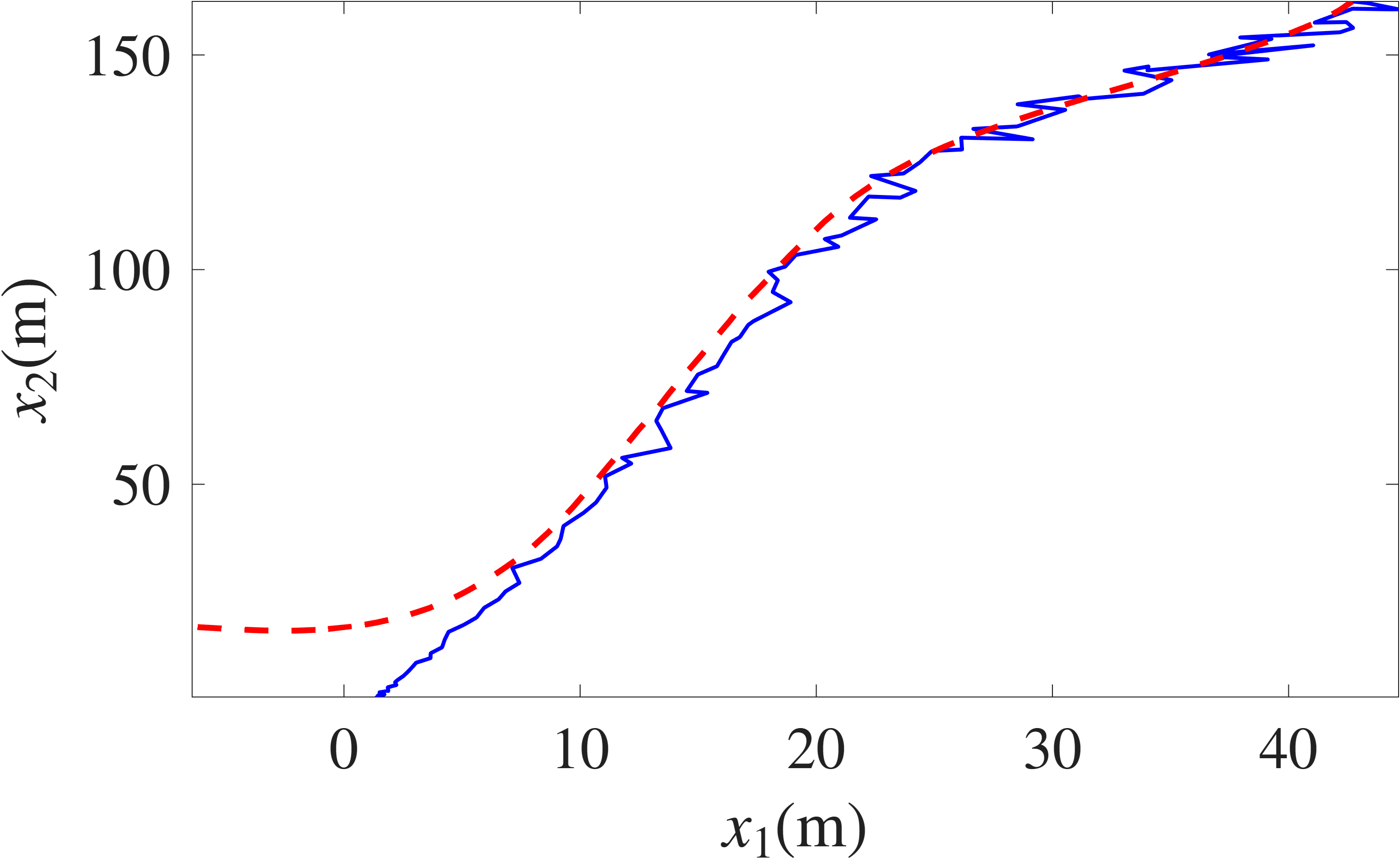}}
		\hspace{\thisfigspace}
		\subfigure{\includegraphics[width=\thisfigwidth]{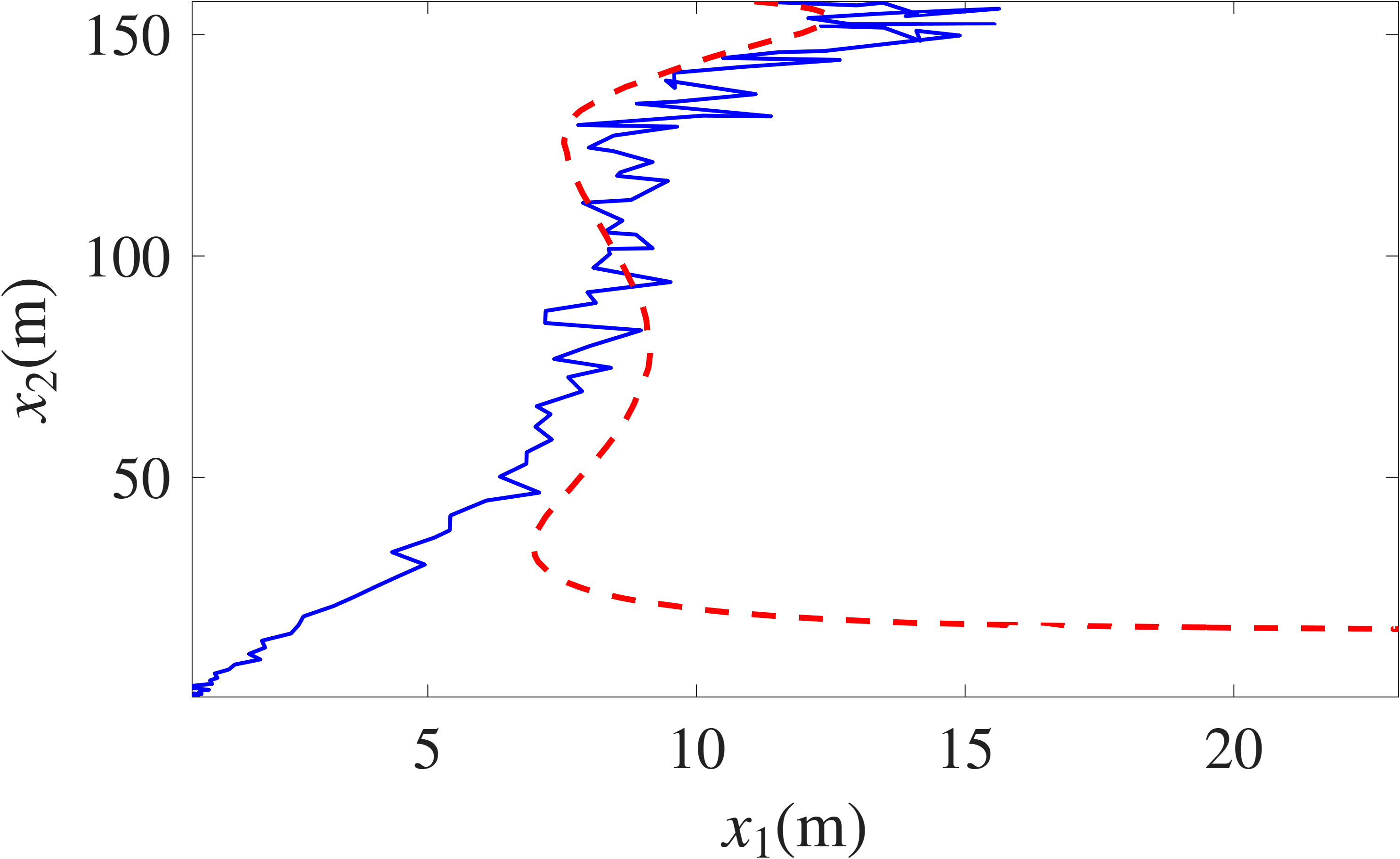}}
		\hspace{\thisfigspace}
	\end{subfigmatrix}
	\caption{ Sample outputs of the S-VAE for 25 real-world samples}
	\label{fig-vae-mars-25}
\end{figure}

\def\thisfigwidth{0.46\columnwidth}
\def\thisfigspace{0.01\columnwidth}
\begin{figure}
	\centering
	\begin{subfigmatrix}{2}
		\subfigure{\includegraphics[width=\thisfigwidth]{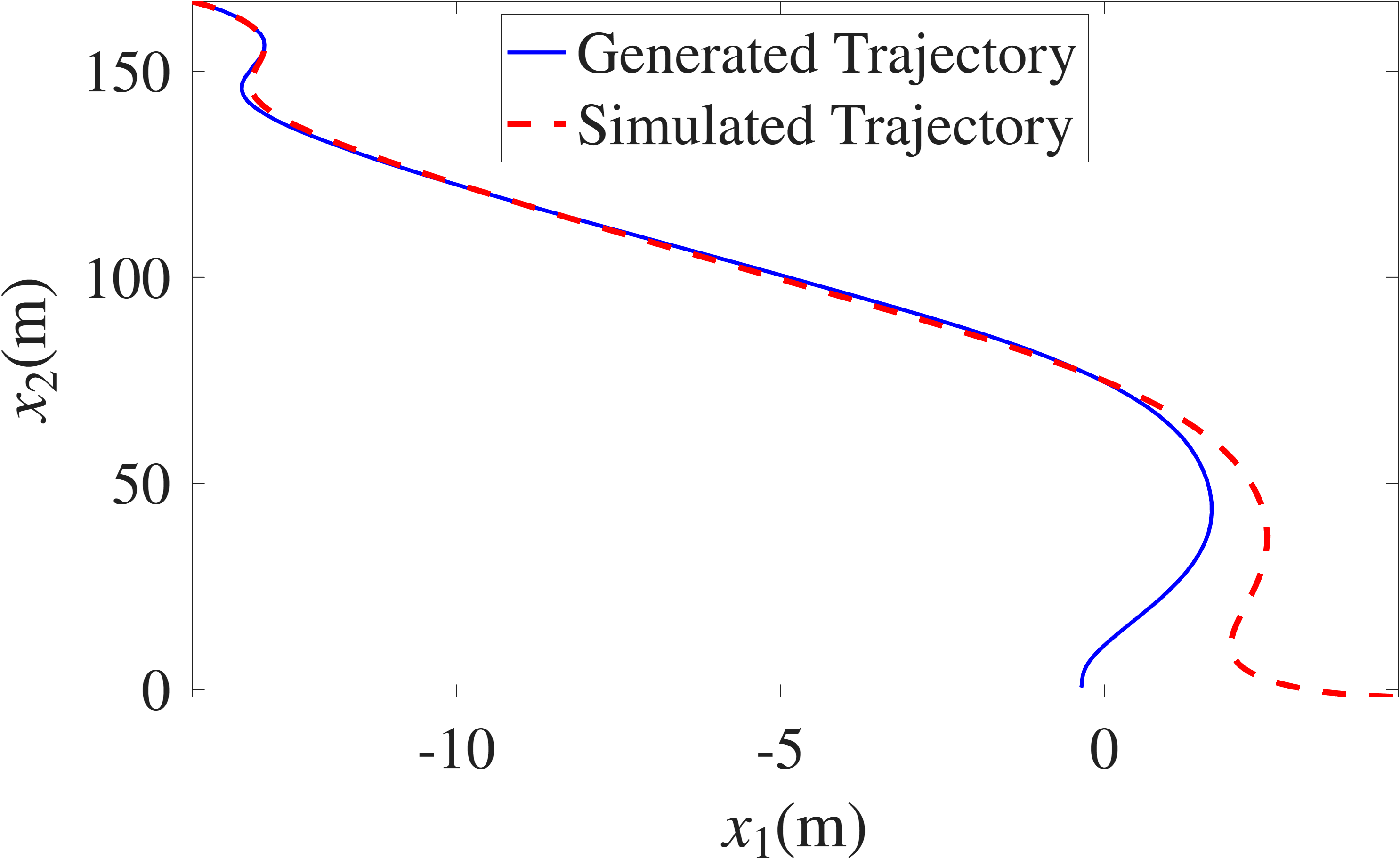}}
		\hspace{\thisfigspace}
		\subfigure{\includegraphics[width=\thisfigwidth]{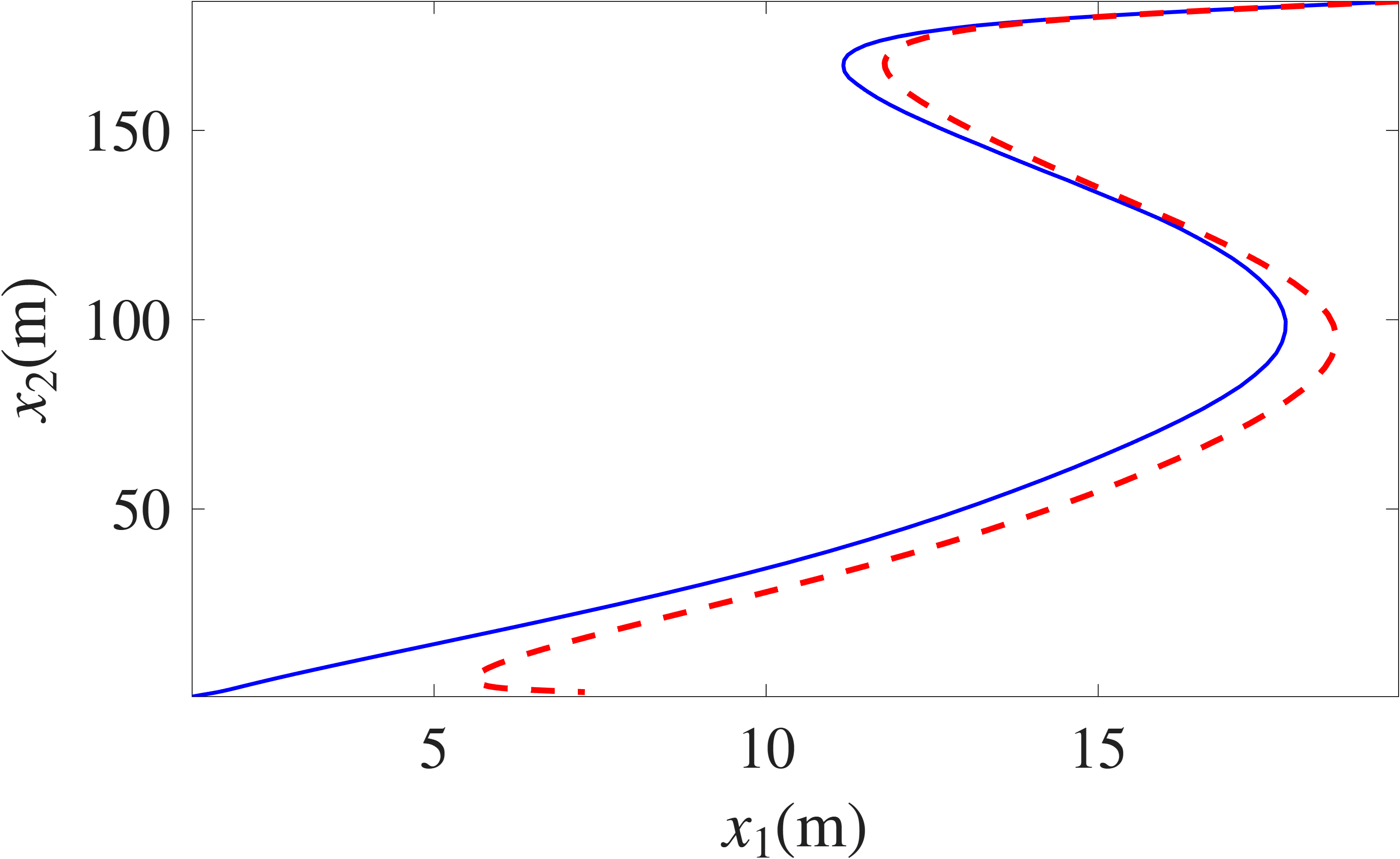}}
		\hspace{\thisfigspace}
		\subfigure{\includegraphics[width=\thisfigwidth]{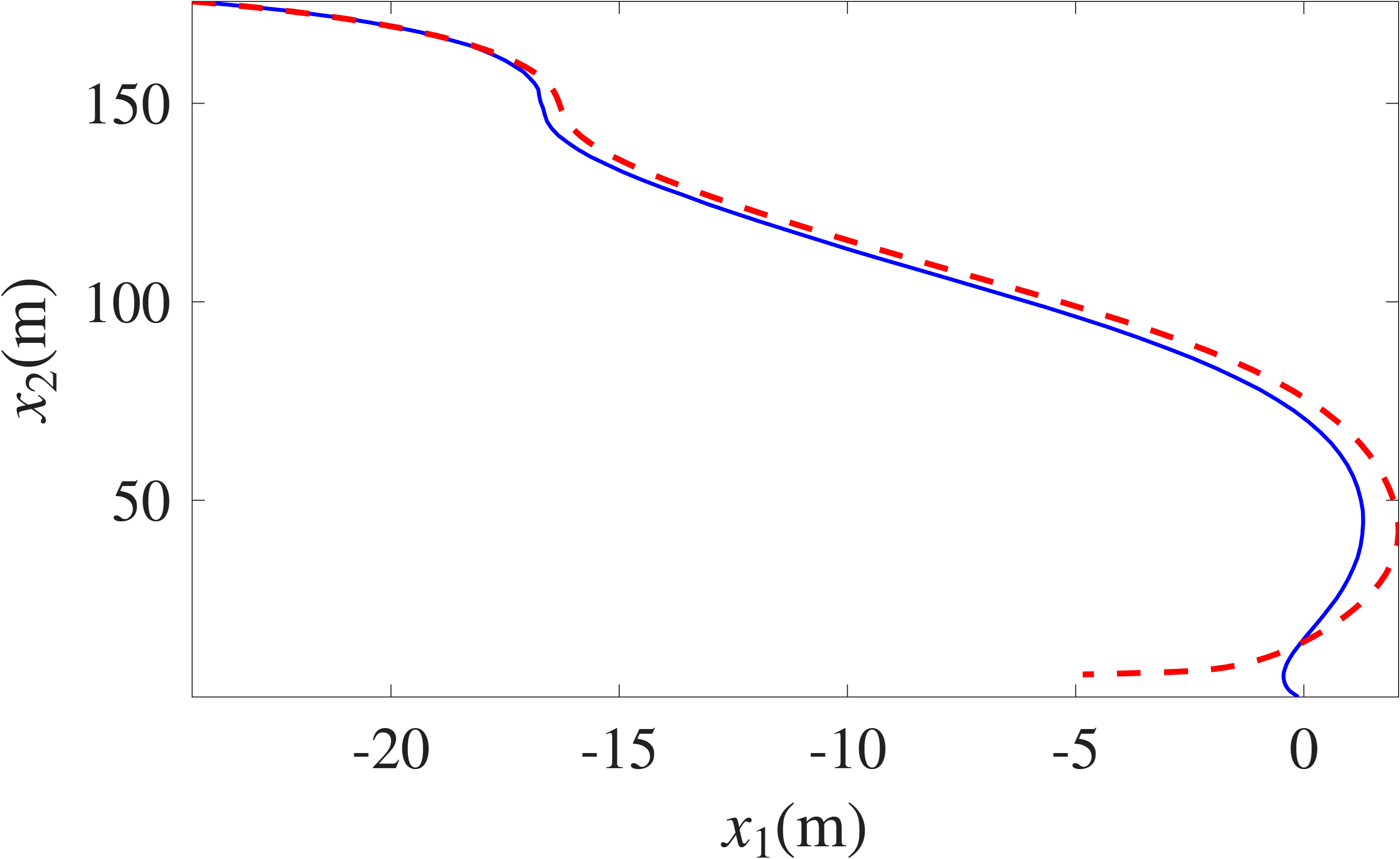}}
		\hspace{\thisfigspace}
		\subfigure{\includegraphics[width=\thisfigwidth]{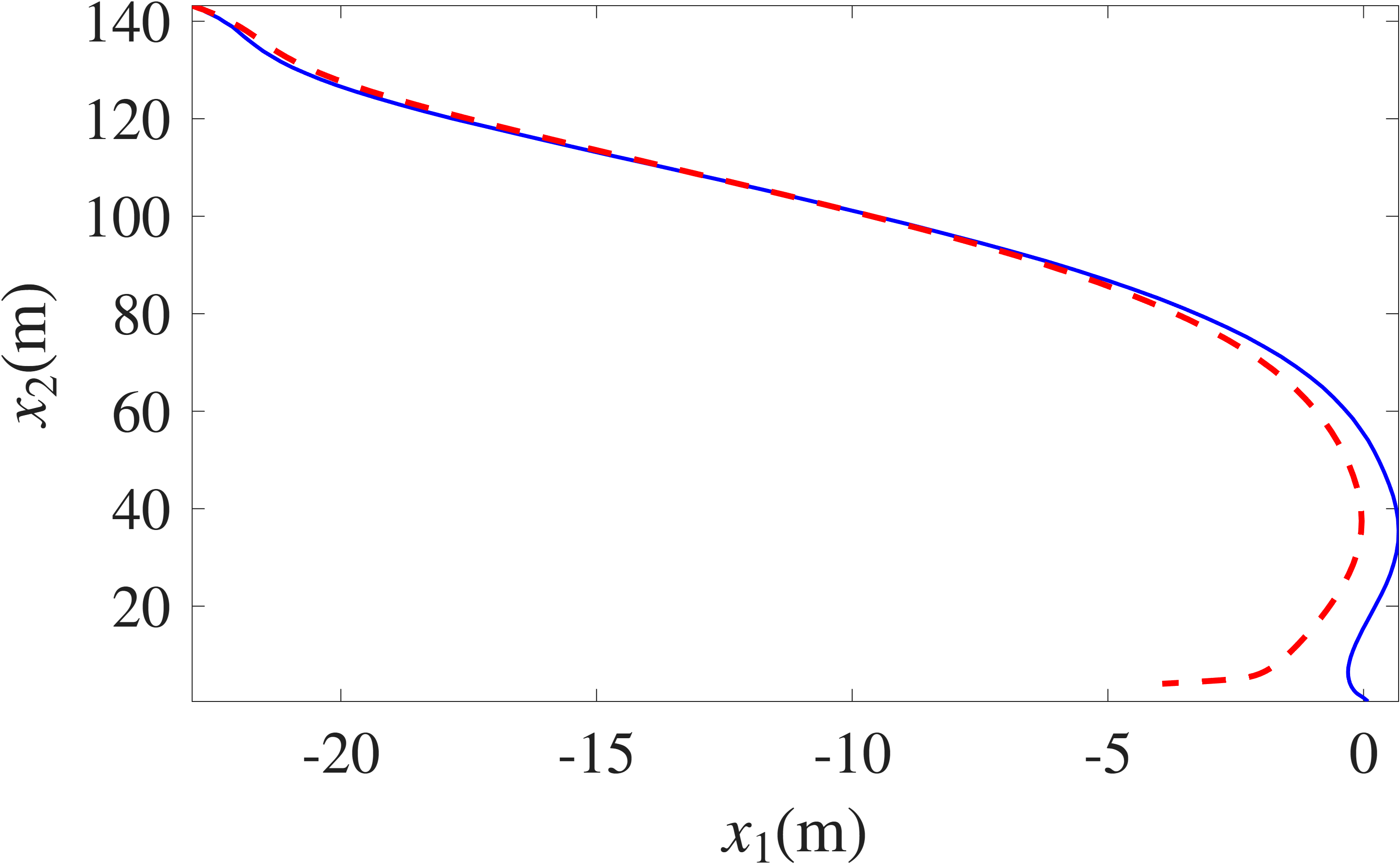}}
		\hspace{\thisfigspace}
	\end{subfigmatrix}
	\caption{ Sample outputs of the S-VAE for 1000 real-world samples}
	\label{fig-vae-mars-1000}
\end{figure}


\paragraph{Statistical Similarity:}
We analyze the first four statistical moments to characterize the data generated by the two neural network models and the OTD. This analysis is performed through a PCA-based projection of both the generated data and the OTD onto the first three principal components.  

In Table~\ref{tbl-vae-marslander-25}, corresponding to 25 $\datasetReal$ samples, we observe that the first two statistical moments of the MI-VAE-generated data closely match those of the OTD, whereas the S-VAE-generated data does not. This agreement is also evident in the PCA scatter plots shown in Figure~\ref{fig-pca-vae-mars-25}. The remaining two moments are comparable to those of the OTD.  
For 1000 $\datasetReal$ samples, Table~\ref{tbl-vae-marslander-1000} shows that both generative models achieve similar performance across all four moments, as further confirmed by the scatter plots in Figure~\ref{fig-pca-vae-mars-1000}.
\def\thisfigwidth{0.46\columnwidth}
\def\thisfigspace{0.01\columnwidth}
\begin{figure}[H]
	\centering
	\subfigure[3-D scatter plot]{\includegraphics[width=\thisfigwidth]{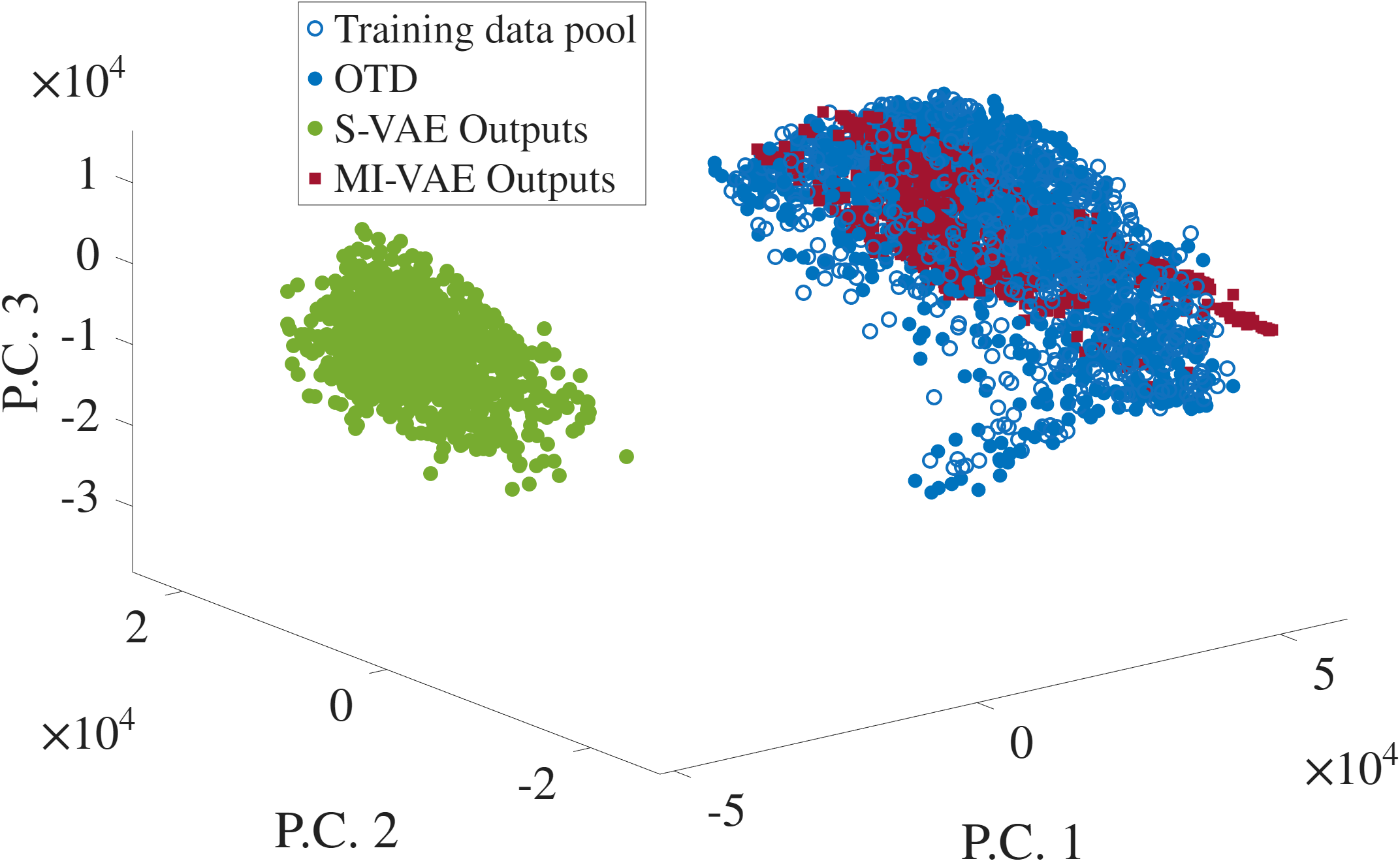}}
	\hspace{\thisfigspace}
	\subfigure[Another view 
	]{\includegraphics[width=\thisfigwidth]{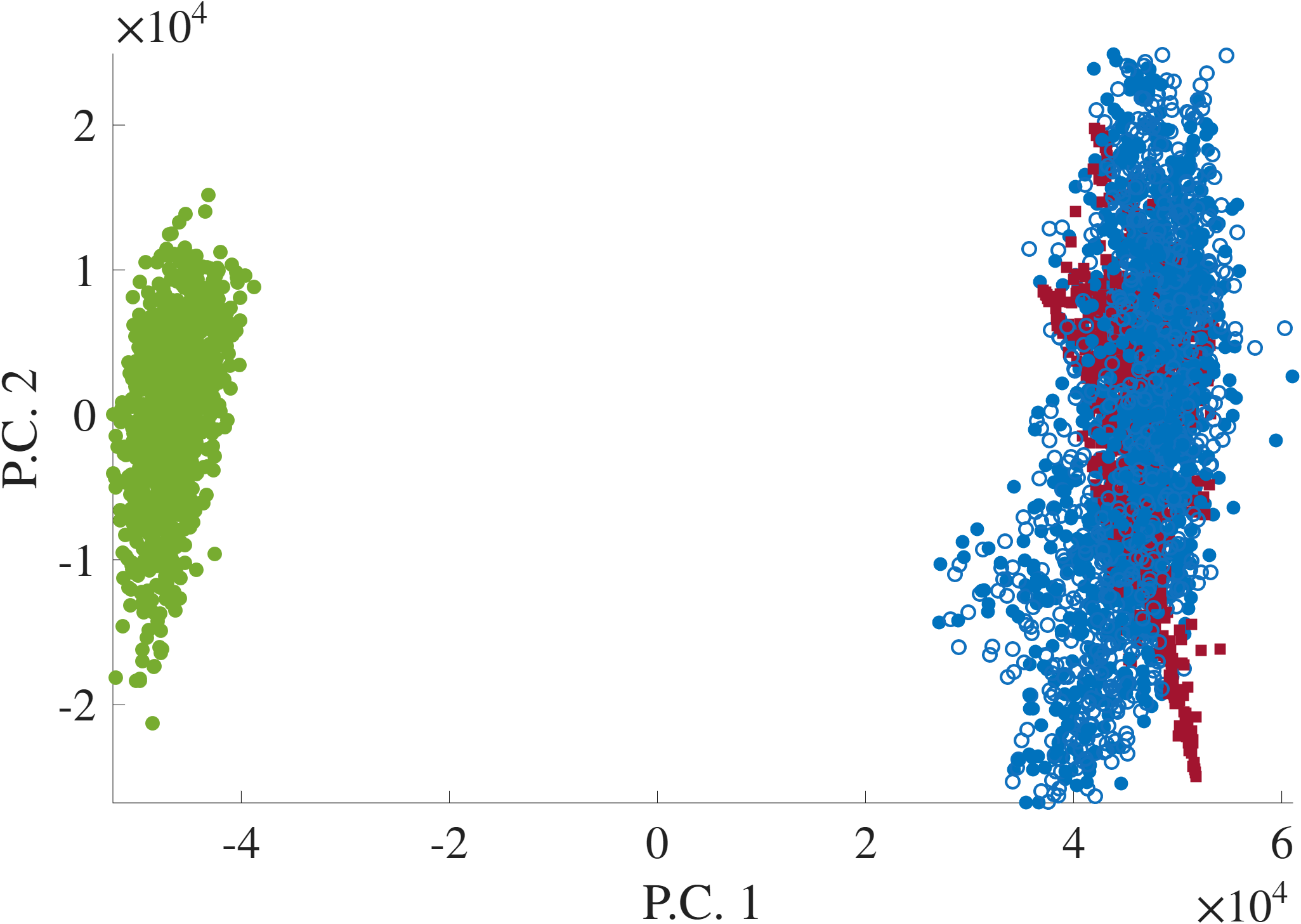}}
	\hspace{\thisfigspace}
	\caption{Scatter plot of first three principal components (P.C.) of data points in the OTD 
		and generated datasets for S-VAE and MI-VAE with $\nData = 25$ and $\nGen = 1000$.}
	\label{fig-pca-vae-mars-25}
\end{figure}

\def\thisfigwidth{0.46\columnwidth}
\def\thisfigspace{0.01\columnwidth}
\begin{figure}[H]
	\centering
	\subfigure[3-D scatter plot]{\includegraphics[width=\thisfigwidth]{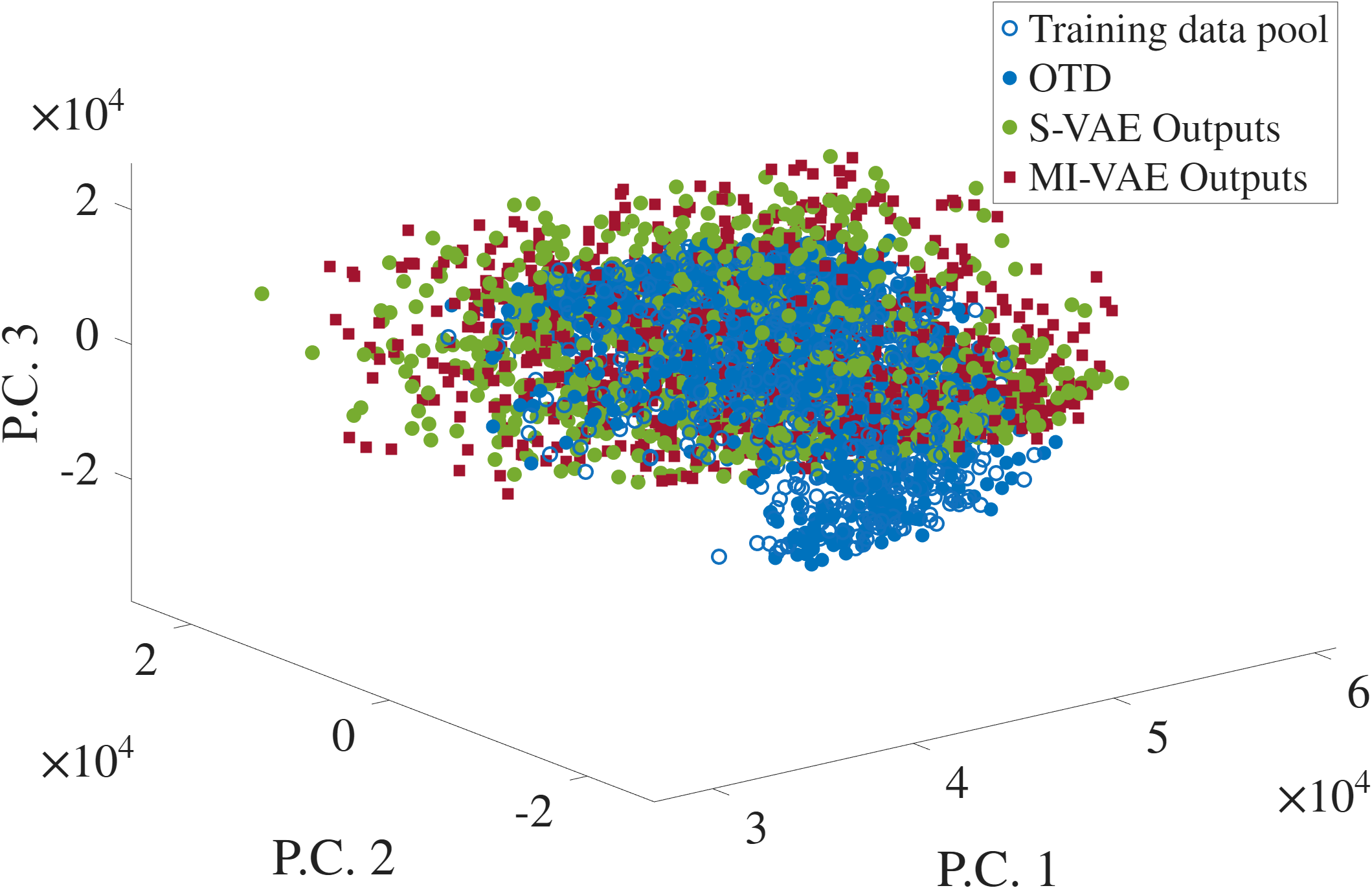}}
	\hspace{\thisfigspace}
	\subfigure[Another view 
	]{\includegraphics[width=\thisfigwidth]{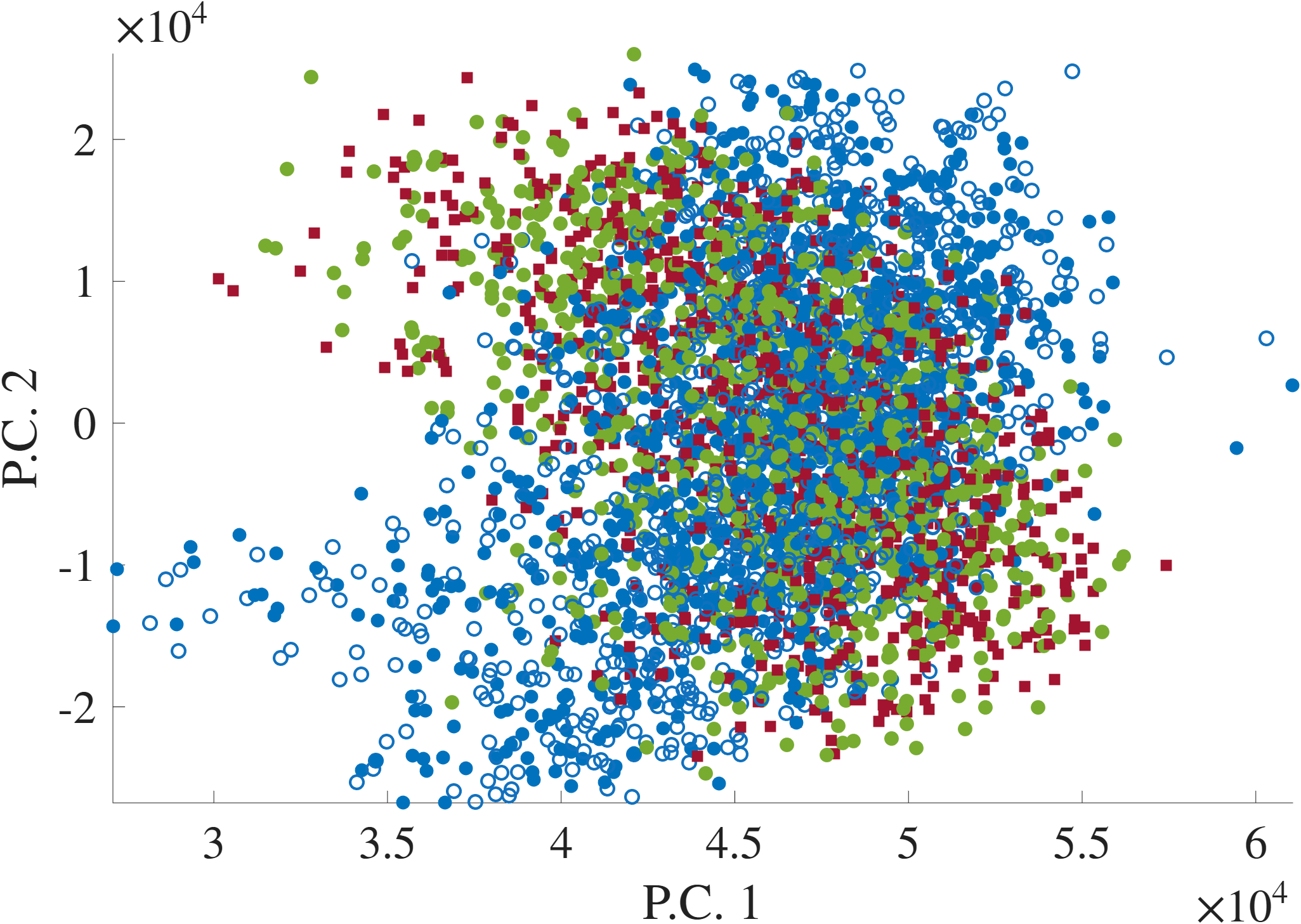}}
	\hspace{\thisfigspace}
	\caption{Scatter plot of first three principal components (P.C.) of data points in the OTD 
		and generated datasets for S-VAE and MI-VAE with $\nData = 1000$ and $\nGen = 1000$.}
	\label{fig-pca-vae-mars-1000}
\end{figure}

\begin{table}[H]
	\centering
	\caption{Statistical moments of datasets generated by the VAE models for $\nData = 25$}
	\label{tbl-vae-marslander-25}
	\resizebox{\textwidth}{!}{%
		\begin{tabular}{lccc|ccc|ccc|ccc}
			\toprule
			\textbf{} & \multicolumn{3}{c|}{Mean} & \multicolumn{3}{c|}{Variance} & \multicolumn{3}{c|}{Skewness} & \multicolumn{3}{c}{Kurtosis} \\
			\midrule
			OTD & $4.598\engE{4}$ & $-5.427\engE{2}$ & $2.142\engE{2}$ & $2.165\engE{7}$ & $1.401\engE{8}$ & $7.319\engE{7}$ & $-0.6578$ & $-0.05233$ & $-1.086$ & $3.752$ & $2.173$ & $4.831$ \\
			S-VAE & $-4.650\engE{4}$ & $1.247\engE{2}$ & $1.440\engE{2}$ & $5.573\engE{6}$ & $3.721\engE{7}$ & $1.779\engE{7}$ & $0.1665$ & $-0.4627$ & $-0.1275$ & $2.802$ & $2.936$ & $2.468$ \\
			MI-VAE & $4.640\engE{4}$ & $2.310\engE{2}$ & $3.101\engE{2}$ & $1.108\engE{7}$ & $8.290\engE{7}$ & $3.460\engE{7}$ & $-0.2651$ & $-0.6804$ & $-0.1179$ & $2.662$ & $3.109$ & $1.970$ \\
			\bottomrule
		\end{tabular}
	}
\end{table}

\begin{table}[H]
	\centering
	\caption{Statistical moments of datasets generated by the VAE models for $\nData = 1000$.}
	\label{tbl-vae-marslander-1000}
	\resizebox{\textwidth}{!}{%
		\begin{tabular}{lccc|ccc|ccc|ccc}
			\toprule
			\textbf{} & \multicolumn{3}{c|}{Mean} & \multicolumn{3}{c|}{Variance} & \multicolumn{3}{c|}{Skewness} & \multicolumn{3}{c}{Kurtosis} \\
			\midrule
			OTD & $4.598\engE{4}$ & $-5.427\engE{2}$ & $2.142\engE{2}$ & $2.165\engE{7}$ & $1.401\engE{8}$ & $7.319\engE{7}$ & $-0.6578$ & $-0.05233$ & $-1.086$ & $3.752$ & $2.173$ & $4.831$ \\
			S-VAE & $4.583\engE{4}$ & $4.464\engE{2}$ & $-1.223\engE{2}$ & $1.901\engE{7}$ & $1.044\engE{8}$ & $6.786\engE{7}$ & $-0.4046$ & $-0.09715$ & $0.7414$ & $2.8917$ & $2.222$ & $3.160$ \\
			MI-VAE & $4.628\engE{4}$ & $4.858\engE{2}$ & $-1.765\engE{2}$ & $1.954\engE{7}$ & $1.029\engE{8}$ & $6.055\engE{7}$ & $-0.4377$ & $-0.07340$ & $0.7093$ & $3.011$ & $2.221$ & $3.342$ \\
			\bottomrule
		\end{tabular}
	}
\end{table}

\subsection{BC and BPPO Results}
\begin{figure}[!htbp]
	\centering
	\begin{tikzpicture}[
		node distance=1mm and 30mm,
		box/.style={draw, rounded corners, align=center, minimum width=3.2cm, minimum height=7mm, fill=white},
		group/.style={draw, dashed, rounded corners, inner sep=6pt, fill=black!3},
		arrow/.style={->, thick}
		]
		
		\node[box] (g1a) {PA-25};
		\node[box, below=of g1a] (g1b) {PA-25 + PB-1000};
		\node[box, below=of g1b] (g1c) {PA-1000};
		
		\begin{scope}[on background layer]
			\node[group, fit=(g1a)(g1b)(g1c),
			label=above:{\textbf{RL Trajectories}}] {};
		\end{scope}
		
		\node[box, right=45mm of g1a] (o1) {RL-25};
		\node[box, right=45mm of g1b] (o2) {RL-Hybrid-25};
		\node[box, right=45mm of g1c] (o3) {RL-1000};
		
		\node[box, below=12mm of g1c] (g2a) {PA-25};
		\node[box, below=of g2a] (g2b) {PA-1000};
		
		\node[box, right=5mm of g2a, text width=15mm, minimum width=1mm](VAE-25){S-VAE};
		\node[box, right=5mm of g2b, text width=15mm, minimum width=1mm](VAE-1000){S-VAE};
		
		\begin{scope}[on background layer]
			\node[group, fit=(g2a)(g2b)(VAE-25)(VAE-1000),
			label=above:{\textbf{S-VAE Trajectories}}] {};
		\end{scope}
		
		\node[box, right=22.5mm of VAE-25] (o4) {VAE-25};
		\node[box, right=22.5mm of VAE-1000] (o5) {VAE-1000};
		
		\node[box, left=12mm of g1c, yshift=-5mm, text width=20mm, minimum width=1mm] (A) {RL Trained on Parameter Set A (PA)};
		\node[box, below=20mm of A, text width=20mm, minimum width=1mm] (B) {RL Trained on Parameter Set B (PB)};
		
		\node[box, below=12mm of g2b] (g3a) {PA-25 + PB-1000};
		\node[box, below=of g3a] (g3b) {PA-1000 + PB-1000};
		
		\node[box, right=5mm of g3a, text width=15mm, minimum width=1mm](MI-25){MI-VAE};
		\node[box, right=5mm of g3b, text width=15mm, minimum width=1mm](MI-1000){MI-VAE};
		
		\begin{scope}[on background layer]
			\node[group, fit=(g3a)(g3b)(MI-25)(MI-1000),
			label=above:{\textbf{MI-VAE Trajectories}}] {};
		\end{scope}
		
		\node[box, right=22.5mm of MI-25] (o6) {MI-VAE-25};
		\node[box, right=22.5mm of MI-1000] (o7) {MI-VAE-1000};
		
		\begin{scope}[on background layer]
			\node[group, fit=(o1)(o2)(o3)(o4)(o5)(o6)(o7),
			label=above:{\textbf{BC and BPPO}}] {};
		\end{scope}
		
		\foreach \n in {g1a,g1b,g1c,g2a,g2b,g3a,g3b}
		\draw[arrow] (A.east) -- (\n.west);
		
		\foreach \n in {g1b,g3a,g3b}
		\draw[arrow] (B.east) -- (\n.west);
		
		\foreach \g/\o/\t in {g1a/o1/25 samples, g1b/o2/1025 samples, g1c/o3/1000 samples, VAE-25/o4/1000 samples, VAE-1000/o5/1000 samples, MI-25/o6/1000 samples, MI-1000/o7/1000 samples}
		\draw[arrow] (\g) -- node[midway, above, font=\footnotesize]{\t} (\o);
		
		\foreach \g/\o in {g2a/VAE-25, g2b/VAE-1000, g3a/MI-25, g3b/MI-1000}
		\draw[arrow] (\g) -- (\o);
		
	\end{tikzpicture}
	
	\caption{Flow of training data to generate BC and BPPO policies.}
	\label{fig:bc-bppo-names}
\end{figure}
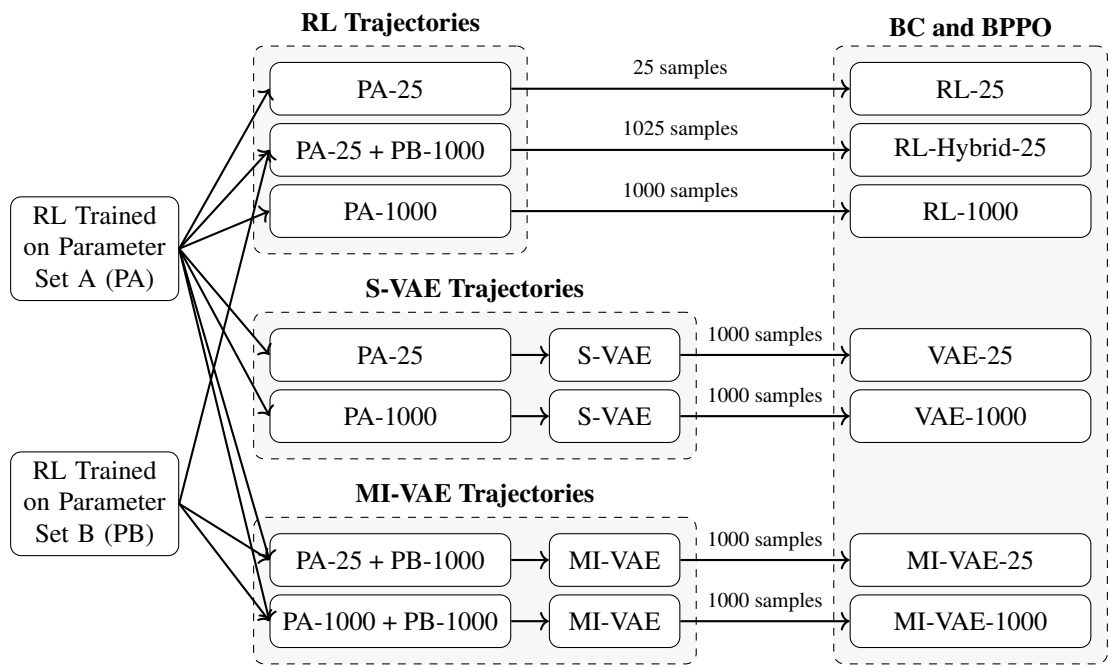

We consider the performance of a BC-trained algorithm and BPPO-trained algorithm over the original dataset and the datasets generated by the S-VAE and MI-VAE models. Figure~\ref{fig:bc-bppo-names} shows the training data used as a basis for training BC and BPPO. The first column specifies sample trajectories generated using an online RL algorithm trained under PA and PB. The middle column specifies the source of the data. The header ``RL Trajectories'' indicates the samples are generated directly from the base RL policy and forward simulated using RK4. The header ``S-VAE Trajectories'' indicates the training samples are generated from the S-VAE model, trained from 25 samples or 1000 samples from PA. The header ``MI-VAE Trajectories'' indicates the training samples are generated from the MI-VAE model, trained from 25 samples or 1000 from PA and 1000 samples from PB. The rightmost set of arrows gives the number of samples used as input to BC and BPPO. The final column gives the associated name of the BC and BPPO models. Note that the BC and BPPO results provided are using the best models generated during training, i.e. the models that achieved the highest reward in a test environment. All BC and BPPO models across all datasets are trained using the same hyperparameters and same seed.   

\def\thisfigwidth{0.46\columnwidth}
\def\thisfigspace{0.01\columnwidth}
\begin{figure}
	\centering
	\begin{subfigmatrix}{2}
		\subfigure{\includegraphics[width=\thisfigwidth]{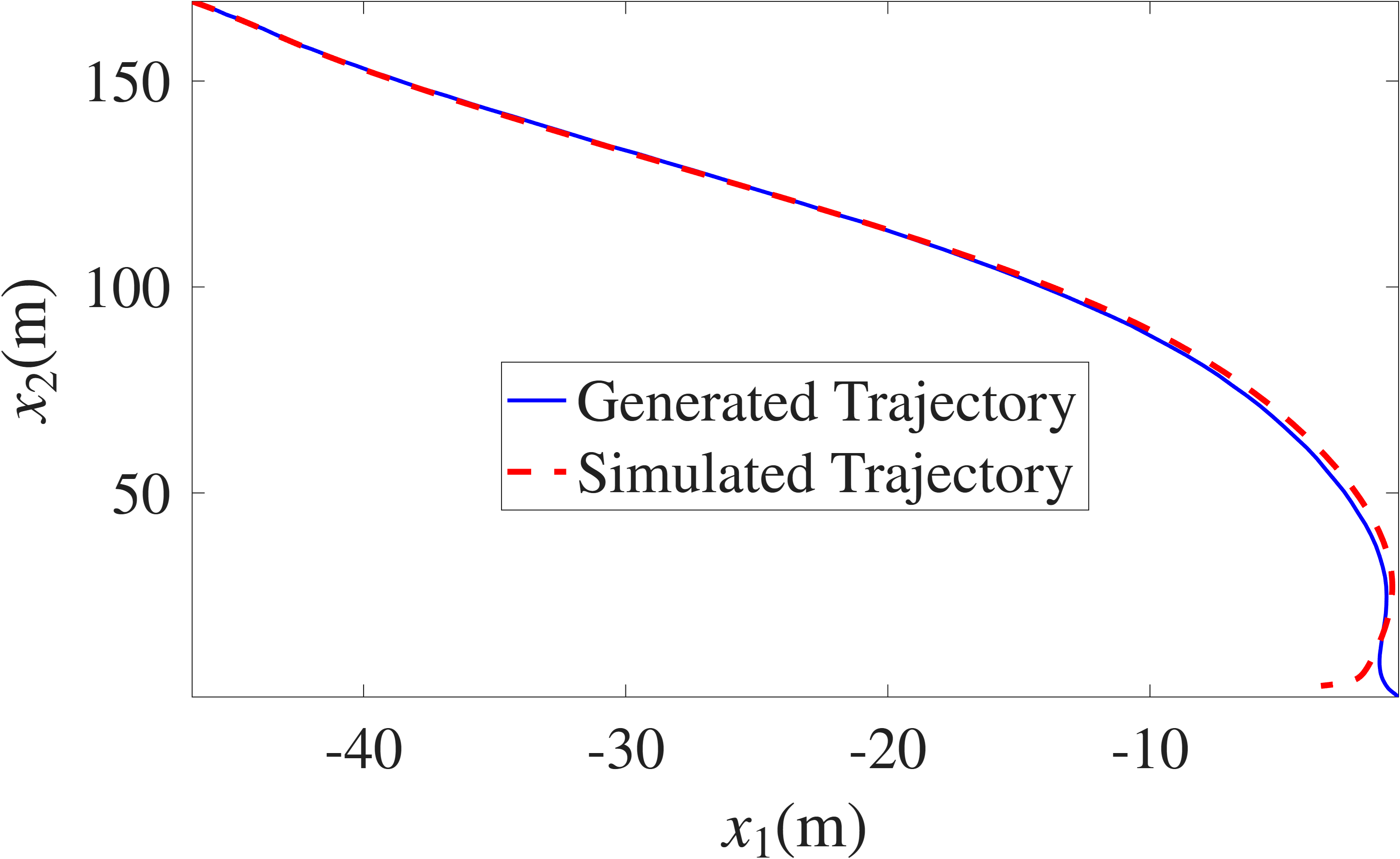}}
		\hspace{\thisfigspace}
		\subfigure{\includegraphics[width=\thisfigwidth]{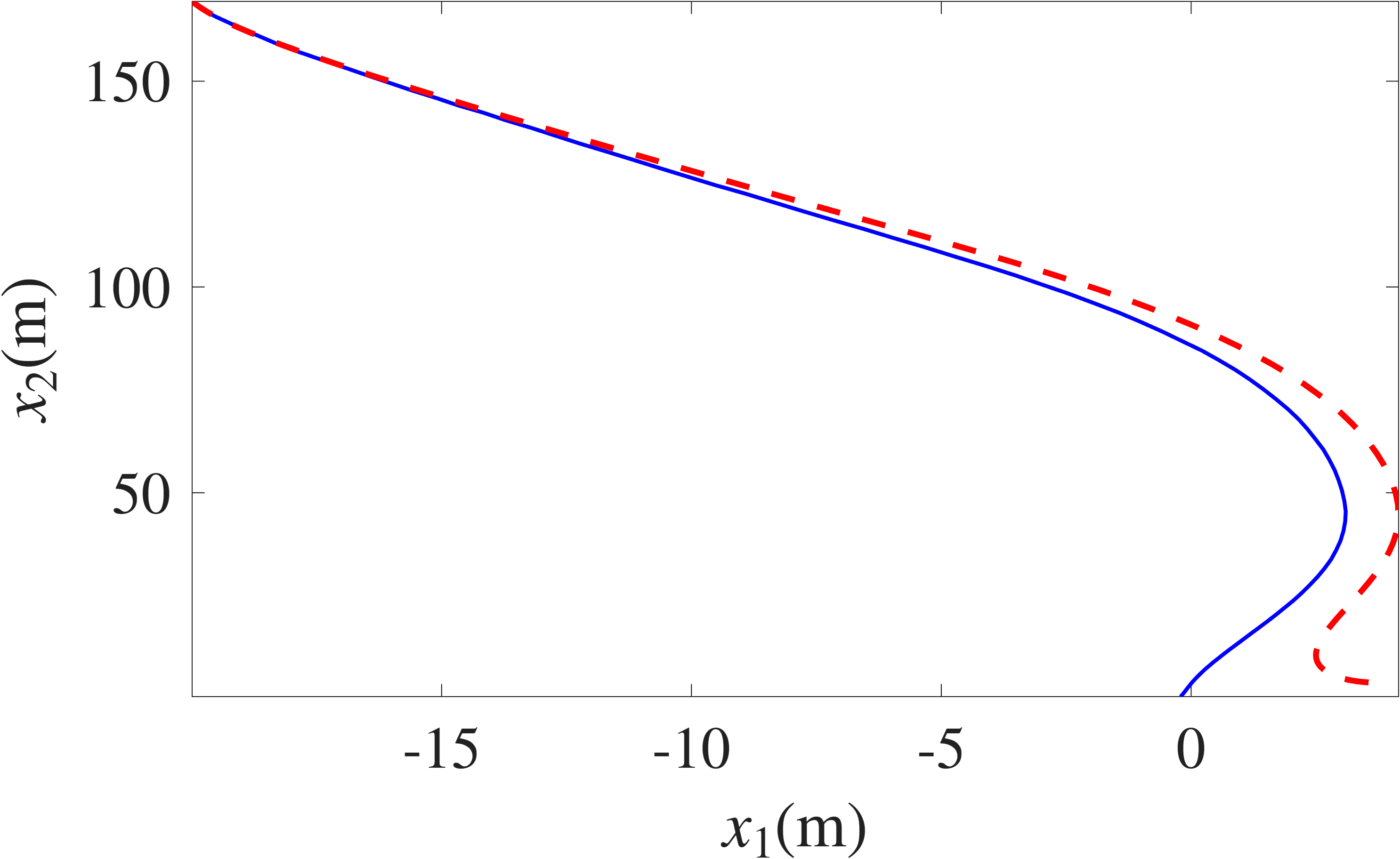}}
		\hspace{\thisfigspace}
		\subfigure{\includegraphics[width=\thisfigwidth]{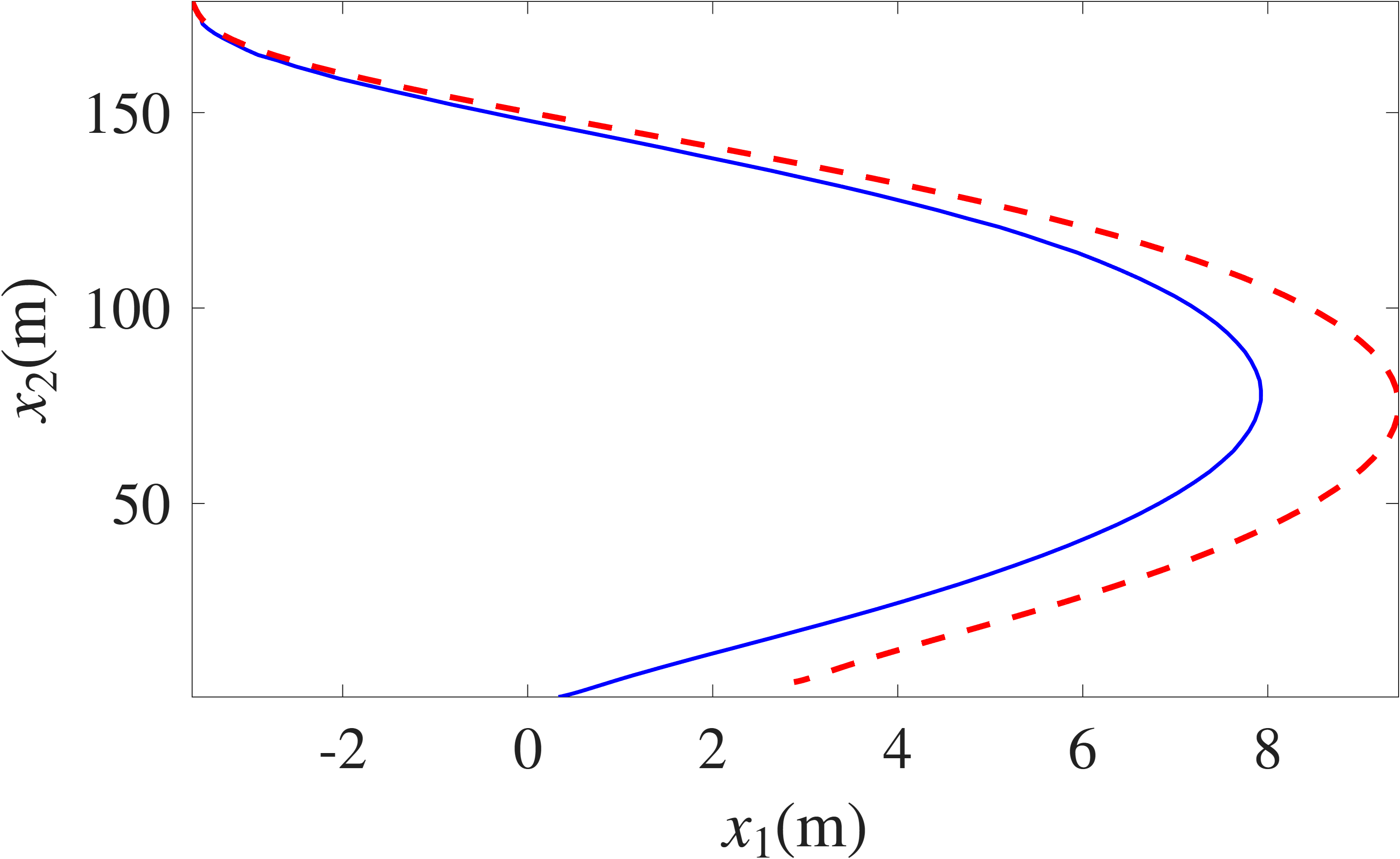}}
		\hspace{\thisfigspace}
		\subfigure{\includegraphics[width=\thisfigwidth]{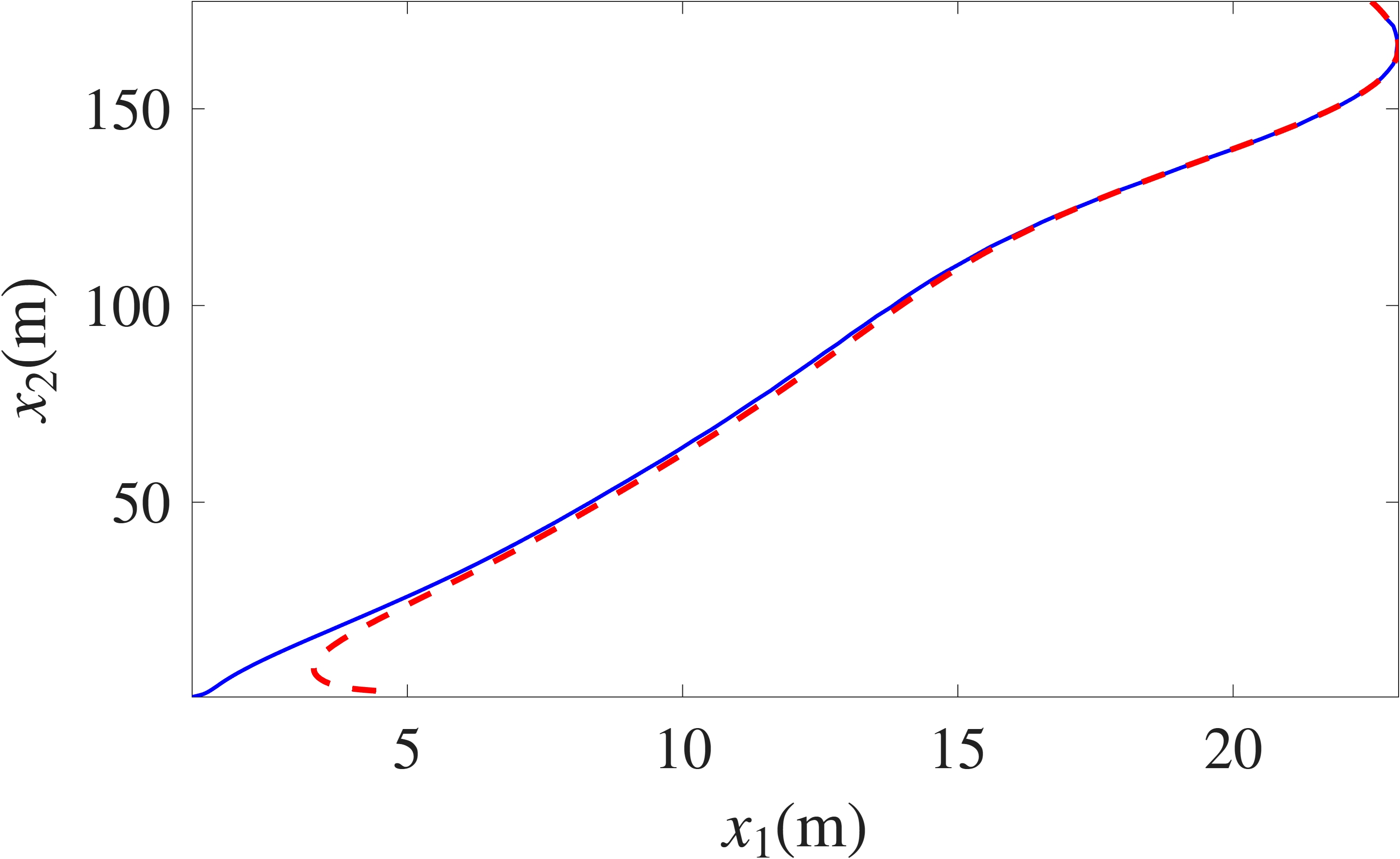}}
		\hspace{\thisfigspace}
	\end{subfigmatrix}
	\caption{ Sample outputs of the MI-VAE for 25 real-world samples}
	\label{fig-MI-VAE-mars-25}
\end{figure}

\def\thisfigwidth{0.46\columnwidth}
\def\thisfigspace{0.01\columnwidth}
\begin{figure}
	\centering
	\begin{subfigmatrix}{2}
		\subfigure{\includegraphics[width=\thisfigwidth]{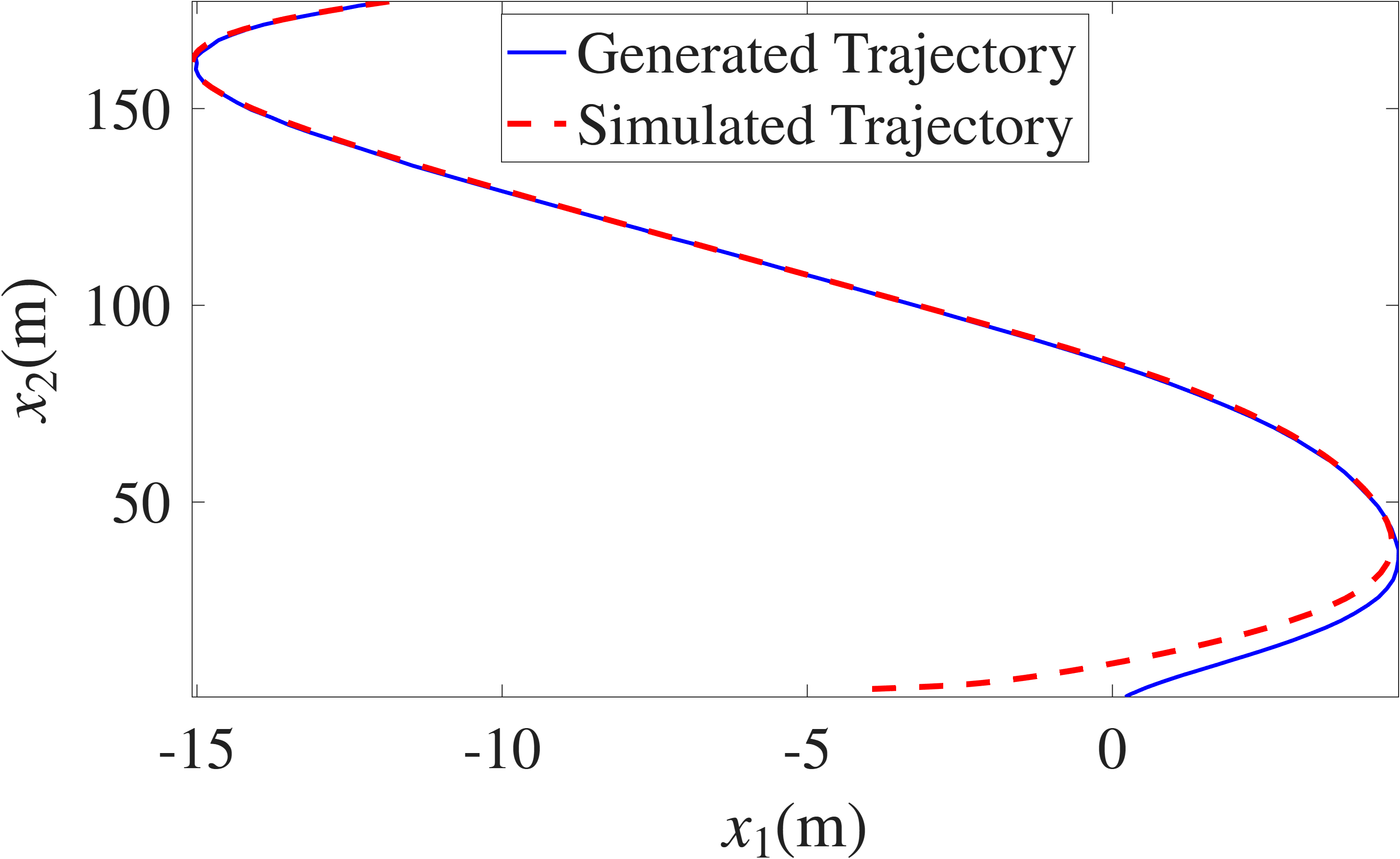}}
		\hspace{\thisfigspace}
		\subfigure{\includegraphics[width=\thisfigwidth]{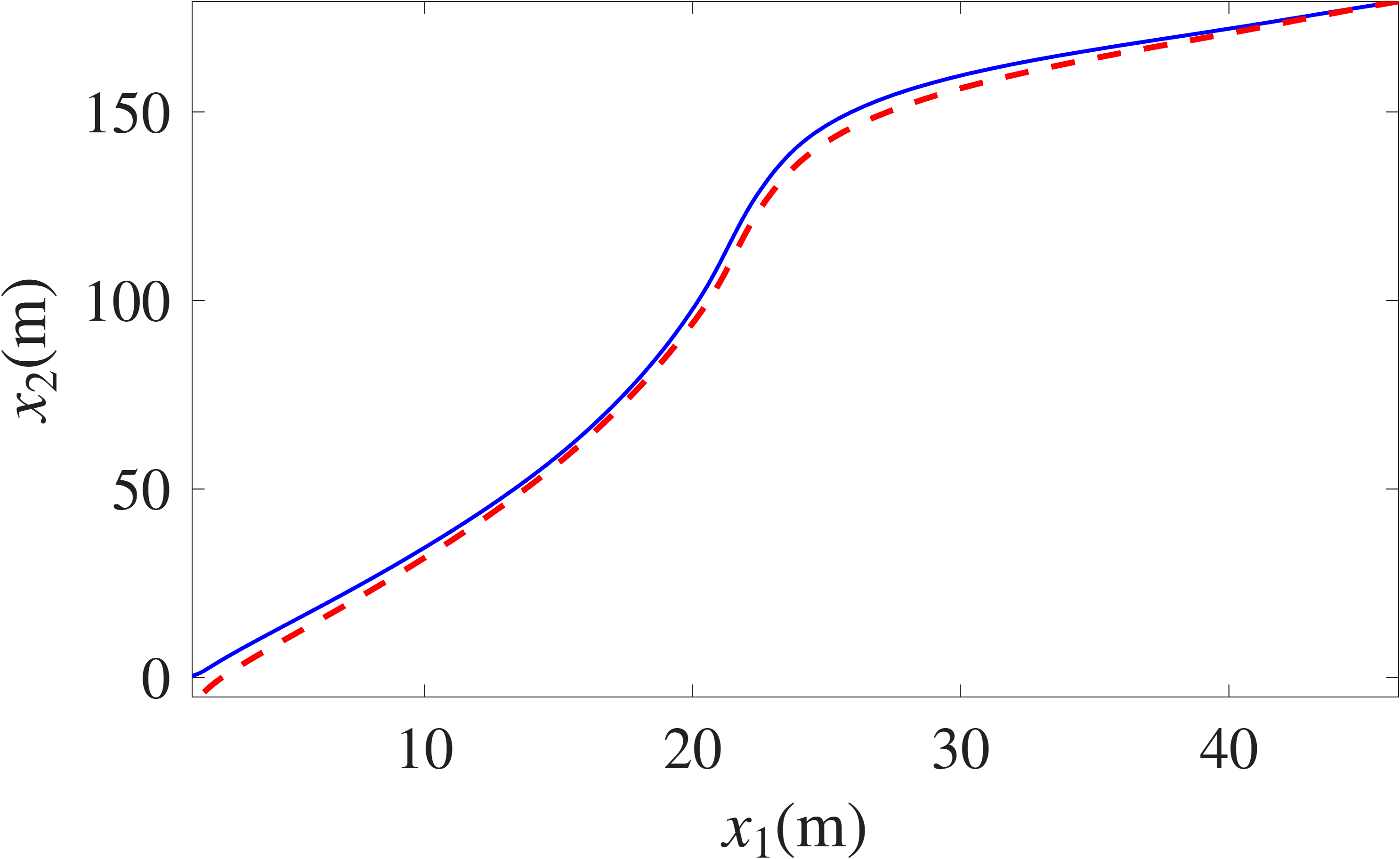}}
		\hspace{\thisfigspace}
		\subfigure{\includegraphics[width=\thisfigwidth]{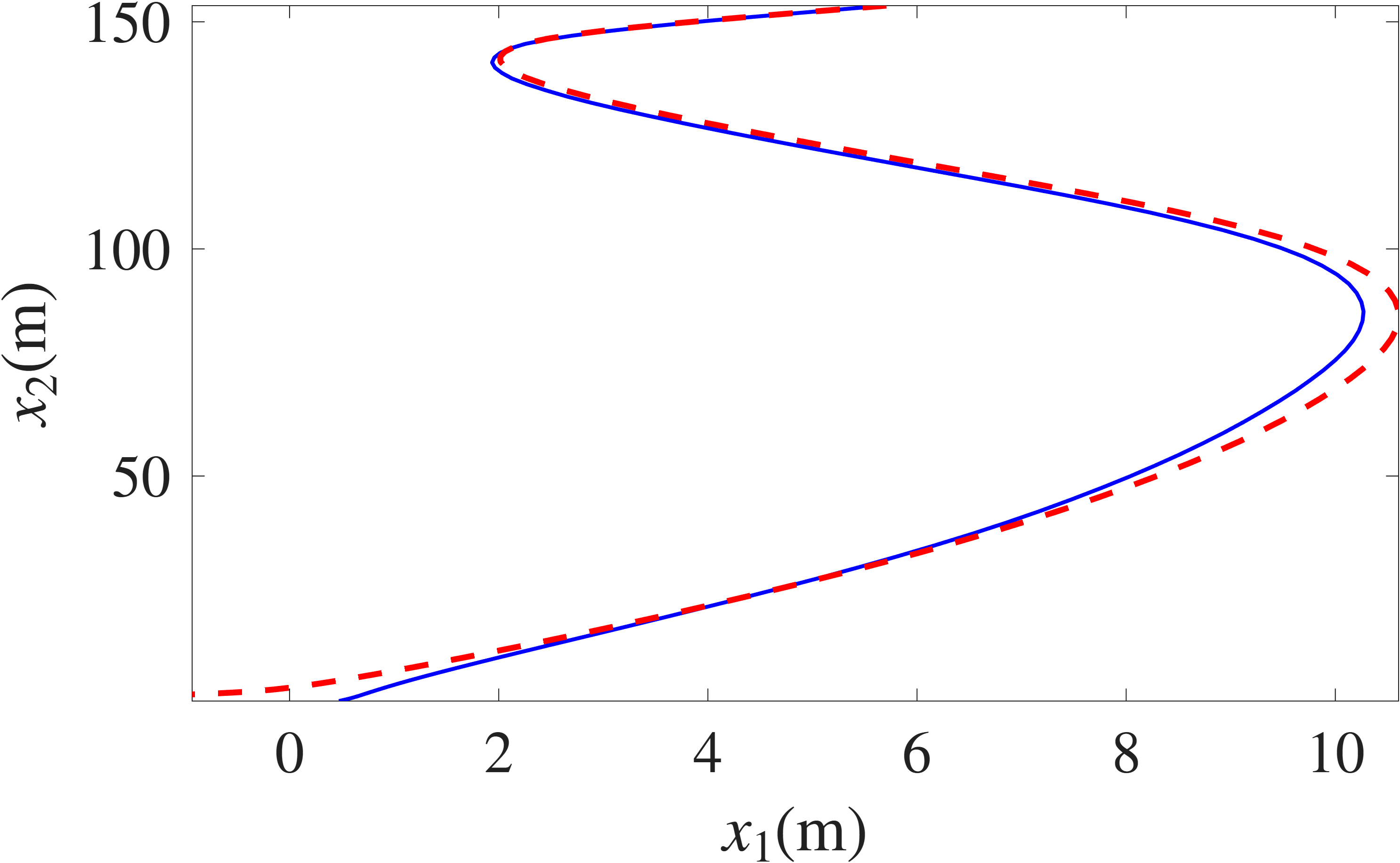}}
		\hspace{\thisfigspace}
		\subfigure{\includegraphics[width=\thisfigwidth]{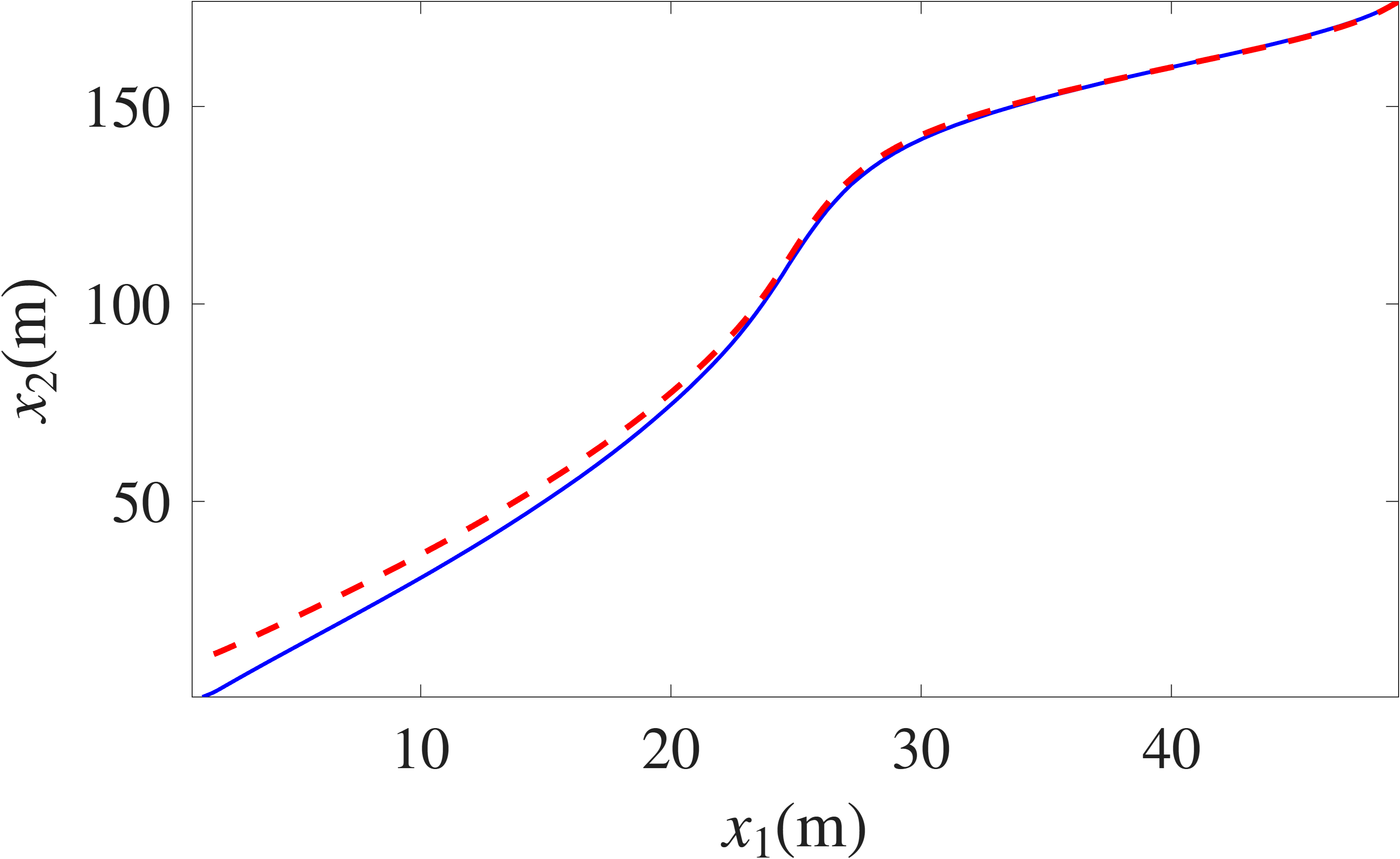}}
		\hspace{\thisfigspace}
	\end{subfigmatrix}
	\caption{ Sample outputs of the MI-VAE for 1000 real-world samples}
	\label{fig-MI-VAE-mars-1000}
\end{figure}

\begin{table*}[!htbp]
\centering
\caption{Average cost and final state deviation for 200 trajectories generated using BC and BPPO policies.}
\label{tab:cost-deviation-policy-results}
\small
\setlength{\tabcolsep}{4pt}  
{\fontsize{9pt}{11pt}
\begin{tabular}{l | c c | c c | c c | c c c }
\hline
\textbf{Dataset}
& \multicolumn{2}{c}{\textbf{Dataset Metrics}}
& \multicolumn{2}{c}{\textbf{BC}}
& \multicolumn{2}{c}{\textbf{BPPO}} \\
\cmidrule(r{5pt}){2-3} \cmidrule(r{5pt}){4-5}\cmidrule(r{5pt}){6-7}
& $\Delta t  \sum| {\yawscalar} |$
& $\sum |\state\msub{f}^\mathrm{err} |  $
& $\Delta t  \sum |{\yawscalar} |$
& $\sum |\state\msub{f}^\mathrm{err} | $
& $\Delta t \sum | \yawscalar |$
& $\sum |\state\msub{f}^\mathrm{err} | $ \\
\hline
RL-25              & $3.99\engE{4} \pm 4.12\engE{3}$ & $4.8 \pm 4.5$ & $3.48\engE{4} \pm 9.54\engE{3}$ & $19.4 \pm 17.0$ & $4.00\engE{4} \pm 9.98\engE{3}$ & $14.4 \pm 16.5$ \\
RL-Hybrid-25 & $5.69\engE{4} \pm 6.21\engE{3}$ & $3.0 \pm 0.9$  & $5.30\engE{4} \pm 2.85\engE{4}$ & $50.0 \pm 27.3$ & $5.37\engE{4} \pm 2.88\engE{4}$ & $50.2 \pm 26.9$ \\
S-VAE-25 &  $3.36\engE{4} \pm 3.57\engE{3}$ & $3.6 \pm 0.6$ & $\mathbf{3.45\engE{4} \pm 5.12\engE{3}}$ & $15.3 \pm 11.7$ & $\mathbf{3.35\engE{4} \pm 5.02\engE{3}}$ & $15.0 \pm 11.1$ \\
MI-VAE-25 & $3.53\engE{4} \pm 5.38\engE{3}$ & $3.8 \pm 0.8$ & $3.57\engE{4} \pm 5.43\engE{3}$ & $\mathbf{10.2 \pm 21.2}$ & $3.69\engE{4} \pm 5.47\engE{3}$ & $\mathbf{7.6 \pm 14.1}$ \\
\midrule
RL-1000 & $3.87\engE{4} \pm 5.95\engE{3}$ & $4.6 \pm 2.0$ & $4.04\engE{4} \pm 5.35\engE{3}$ & $\mathbf{3.4 \pm 2.1}$ & $4.22\engE{4} \pm 5.62\engE{3}$ & $\mathbf{2.8 \pm 1.6}$ \\
S-VAE-1000  & $3.74\engE{4} \pm 4.24\engE{3}$ & $4.2 \pm 1.1$ & $3.82\engE{4} \pm 4.64\engE{3}$ & $3.7 \pm 1.5$ & $\mathbf{3.81\engE{4} \pm 4.58\engE{3}}$ & $3.7 \pm 1.4$ \\
MI-VAE-1000  & $3.70\engE{4} \pm 4.47\engE{3}$ & $4.1 \pm 1.1$ & $\mathbf{3.80\engE{4} \pm 4.40\engE{3}}$ & $4.0 \pm 2.0$ & $3.84\engE{4} \pm 4.38\engE{3}$ & $3.8 \pm 1.7$ \\


\hline
\end{tabular}}
\end{table*}

\paragraph{Interpretation of Results:} Given 25 training samples, the MI-VAE is able to provide the most consistent performance, with the highest reward and success rate over both BC and BPPO. Providing the 25 directly as training data for BPPO also leads to decent performance, but has a lower reward and success rate. The S-VAE data is not able to accurate capture the trajectory information, leading to performance degradation under BPPO. Supplementing the 25 samples from $\datasetReal$ with 1000 samples from $\datasetIdeal$ (as is used for training the MI-VAE model) does not lead to a meaningful reward. 
Given 1000 training samples, we see that the RL-1000 BC and BPPO models perform marginally better than the MI-VAE and S-VAE models, but all three datasets provide sufficient data for good convergence. 

\begin{figure}[H]
	\centering
	
	\begin{minipage}{0.48\textwidth}
		\centering
		\begin{tikzpicture}
			\begin{axis}[
				ybar,
				bar width=7pt,
				width=\linewidth,
				height=6.5cm,
				ylabel={Reward},
				symbolic x coords={RL-25, RL-Hybid-25, S-VAE-25, MI-VAE-25},
				xtick=data,
				xticklabel style={rotate=45, anchor=east, font=\small},
				enlarge x limits=0.25,
				ymin=0, ymax=120, 
				legend style={at={(0.5,-0.35)}, anchor=north, legend columns=-1, font=\tiny},
				error bars/y dir=both,
				error bars/y fixed,
				error bars/error mark options={rotate=90, black, thick, mark size=2pt}
				]
				\addplot+[error bars/.cd, y dir=both, y explicit] 
				coordinates {(RL-25,82.9) +- (0,6.8) (RL-Hybid-25,85.9) +- (0,1.7) (S-VAE-25,87.6) +- (0,2.9) (MI-VAE-25,85.5) +- (0,3.9)};
				\addplot+[error bars/.cd, y dir=both, y explicit] 
				coordinates {(RL-25,80.8) +- (0,17.0) (RL-Hybid-25,46.5) +- (0,30.8) (S-VAE-25,85.0) +- (0,11.7) (MI-VAE-25,90.1) +- (0,21.3)};
				\addplot+[error bars/.cd, y dir=both, y explicit] 
				coordinates {(RL-25,86.0) +- (0,16.6) (RL-Hybid-25,46.3) +- (0,30.4) (S-VAE-25,85.3) +- (0,11.2) (MI-VAE-25,92.8) +- (0,14.1)};
				
				\legend{Dataset, BC, BPPO}
			\end{axis}
		\end{tikzpicture}
		\vspace{1.2cm}
		
	\end{minipage}
	\hfill
	\begin{minipage}{0.48\textwidth}
		\centering
		\begin{tikzpicture}
			\begin{axis}[
				ybar,
				bar width=7pt,
				width=\linewidth,
				height=6.5cm,
				ylabel={Success (\%)},
				symbolic x coords={RL-25, RL-Hybrid-25, S-VAE-25, MI-VAE-25},
				xtick=data,
				xticklabel style={rotate=45, anchor=east, font=\small},
				enlarge x limits=0.25,
				ymin=0, ymax=110,
				legend style={at={(0.5,-0.35)}, anchor=north, legend columns=-1, font=\tiny},
				]
				\addplot coordinates {(RL-25,96) (RL-Hybrid-25,99.9) (S-VAE-25,100) (MI-VAE-25,100)};
				\addplot coordinates {(RL-25,28) (RL-Hybrid-25,2.5) (S-VAE-25,36) (MI-VAE-25,62.5)};
				\addplot coordinates {(RL-25,55.5) (RL-Hybrid-25,2.5) (S-VAE-25,31) (MI-VAE-25,79.5)};
				\legend{Dataset, BC, BPPO}
			\end{axis}
		\end{tikzpicture}
		\vspace{1.2cm}
		
	\end{minipage}
	
	\vspace{0.5cm}
	\caption{Reward with Std. Dev. (T-lines) and Success Rate across datasets for $\nData =25$.}
	\label{fig:metrics-25}
\end{figure}
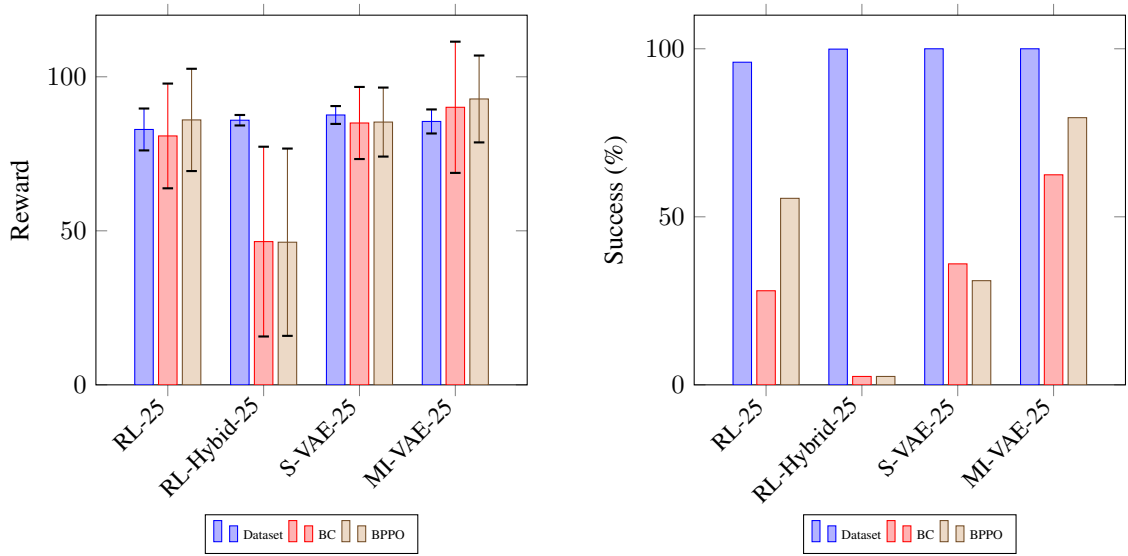

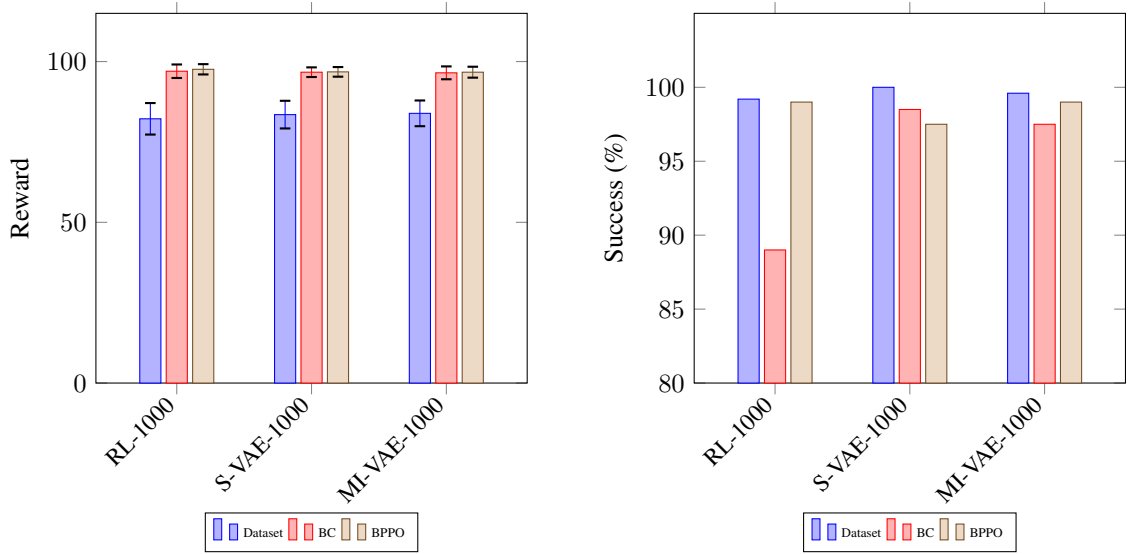
\begin{figure}[H]
	\centering
	
	\begin{minipage}{0.48\textwidth}
		\centering
		\begin{tikzpicture}
			\begin{axis}[
				ybar,
				bar width=8pt,
				width=\linewidth,
				height=6.5cm,
				ylabel={Reward},
				symbolic x coords={RL-1000, S-VAE-1000, MI-VAE-1000},
				xtick=data,
				xticklabel style={rotate=45, anchor=east, font=\small},
				enlarge x limits=0.3,
				ymin=0, ymax=115,
				legend style={at={(0.5,-0.35)}, anchor=north, legend columns=-1, font=\tiny},
				error bars/y dir=both,
				error bars/y fixed,
				error bars/error mark options={rotate=90, black, thick, mark size=2pt}
				]
				\addplot+[error bars/.cd, y dir=both, y explicit] 
				coordinates {(RL-1000,82.2) +- (0,4.9) (S-VAE-1000,83.5) +- (0,4.3) (MI-VAE-1000,83.9) +- (0,4.0)};
				\addplot+[error bars/.cd, y dir=both, y explicit] 
				coordinates {(RL-1000,97.0) +- (0,2.1) (S-VAE-1000,96.7) +- (0,1.5) (MI-VAE-1000,96.5) +- (0,2.0)};
				\addplot+[error bars/.cd, y dir=both, y explicit] 
				coordinates {(RL-1000,97.6) +- (0,1.6) (S-VAE-1000,96.8) +- (0,1.5) (MI-VAE-1000,96.7) +- (0,1.7)};
				
				\legend{Dataset, BC, BPPO}
			\end{axis}
		\end{tikzpicture}
		\vspace{1.2cm}
	\end{minipage}
	\hfill
	\begin{minipage}{0.48\textwidth}
		\centering
		\begin{tikzpicture}
			\begin{axis}[
				ybar,
				bar width=8pt,
				width=\linewidth,
				height=6.5cm,
				ylabel={Success (\%)},
				symbolic x coords={RL-1000, S-VAE-1000, MI-VAE-1000},
				xtick=data,
				xticklabel style={rotate=45, anchor=east, font=\small},
				enlarge x limits=0.3,
				ymin=80, ymax=105, 
				ytick={80,85,90,95,100},
				legend style={at={(0.5,-0.35)}, anchor=north, legend columns=-1, font=\tiny},
				]
				\addplot coordinates {(RL-1000,99.2) (S-VAE-1000,100) (MI-VAE-1000,99.6)};
				\addplot coordinates {(RL-1000,89) (S-VAE-1000,98.5) (MI-VAE-1000,97.5)};
				\addplot coordinates {(RL-1000,99) (S-VAE-1000,97.5) (MI-VAE-1000,99)};
				
				\legend{Dataset, BC, BPPO}
			\end{axis}
		\end{tikzpicture}
		\vspace{1.2cm}
	\end{minipage}
	
	\vspace{0.5cm}
	\caption{Reward with Std. Dev. and Success Rate for  $\nData =1000$.}
	\label{fig:metrics-1000}
\end{figure}

Table~\ref{tab:cost-deviation-policy-results}  and Figures.\ref{fig:metrics-25}--\ref{fig:metrics-1000} give the performance of 200 trajectories generated using the BC and BPPO algorithms trained on these seven datasets. There are two main subgroups: results for models trained on 25 samples from $\datasetReal$ and results for models trained on 1000 samples from $\datasetReal$. The best results for 25 samples and 1000 samples are bolded for each performance metric. Note for BPPO, we modify the reward function from \eqref{eq:reward-ppo}. We remove the shaping term and modify the scaling on the control cost and terminal states, as given in \eqref{eq:reward-bppo}.
\begin{align}
r &= - 0.1 \sum_{i=1}^{3} \frac{\yawscalar_i(t)}{10^3}
     + \indicator_{\{\ypos < 1\}} \Bigg( 100
       - |\xpos|
       - \frac{180}{\policy} |\orient|
       - |\xvel + \windx|
       - |\yvel + \windy| \Bigg) \label{eq:reward-bppo}
\end{align}
With BPPO, we have the ability to remove the shaping term as we are not concerned with the initial convergence of the algorithm. Instead, we want to optimize the overall performance, which we define as the integral control cost and final state deviation, using the simulated trajectories as training data.

There are 4 performance metrics considered: cumulative reward, success rate, cumulative control cost $\Delta t  \sum| {\yawscalar} |$, and final state deviation $ \sum |\state\msub{f}^\mathrm{err} | = \sum_{i=0}^5 |\state\msub{f} - \state\msub{f}^\mathrm{target} |$. The reward is the signal received during training BPPO. It measures the combined optimization of all parameters. The success rate indicates what percent of the time the final state is within the bounds given in Table~\ref{tab:state-bounds}. The control cost is the total control used by the vehicle during landing, and the final state deviation gives the deviation from the target values for all bounded states in Table~\ref{tab:state-bounds}.

Behavior cloning is supervised learning, which learns to map states to actions, i.e. the learned policy mimics the actions taken in the sample dataset. The performance of BC indicates how well the model is able to learn the correlation between states and actions. Good performance under BC means that the dataset covers the state distribution and maps that to meaningful actions.

Behavior proximal policy optimization is an offline RL algorithm, which means that it uses the training dataset to optimize the cumulative reward. BPPO indicates whether the dataset provides meaningful, consistent information about the value of states and state-action pairs. If BPPO provides a performance improvement over BC, the trajectories are consistent and meaningful and allow a good estimation of the relative value of different states and actions.

\section{Conclusion}\label{sec-con}
In this paper, we analyzed a representative example problem to address the data scarcity issue in RL training. To this end, we developed a mutual-information-based variational autoencoder with a split latent space designed to jointly map and disentangle shared and distinct features of the input data. We evaluated the proposed MI-VAE against a standard VAE (S-VAE) under varying training data regimes, specifically for $\nData = 25$ and $\nData = 1000$. The results indicate that the MI-VAE outperforms the S-VAE across the predefined performance metrics when $\nData = 25$, while both models exhibit comparable performance for the larger dataset. These findings suggest that, in data-limited settings, the MI-VAE yields superior training outcomes relative to the standard VAE framework. 

Moreover, this study demonstrates the potential of the MI-VAE for addressing reinforcement learning problems arising from control systems applications. Training an RL algorithm using offline data may be desirable for a number of reasons, such as reduction of samples needed from the environment, better generalization to a real-world system, and mimicking pre-established desired behavior. The results from the BPPO model trained on the MI-VAE indicate promising results in environments with high costs to generate samples, with better performance and generalization than direct use of the samples for $\nData=25$. These results suggest that the use of the MI-VAE enhances the ability of offline RL algorithms when training data is limited. 

\section{Future Scope}\label{sec-scope}
Future work will extend the proposed framework to a broader class of problems within the control systems literature. In particular, we plan to incorporate an explicit sensor model, enabling the system to account for both process and measurement noise, thereby more accurately capturing real-world operating conditions. To further validate the proposed approach, we will conduct experiments using data collected from real-world trials on a physical system. We also intend to present a formal mathematical derivation of the MI-VAE loss function and rigorously establish its theoretical validity. To further strengthen the applicability of the proposed approach to real-world control and reinforcement learning problems, we will be focusing on developing a control-conditioned machine learning framework capable of generating an ensemble of system states conditioned on a given control input. Such a model would explicitly capture the stochasticity and uncertainty inherent in system dynamics, enabling probabilistic prediction of state evolution under different control actions. Finally, future studies will investigate robustness, scalability to higher-dimensional systems, and integration of the learned representations into downstream reinforcement learning and control pipelines.

\section*{Acknowledgements}

\bibliographystyle{AAS_publication}
\bibliography{References}
\end{document}